\documentclass[journal]{IEEEtran}
\usepackage{amsmath,amsfonts}
\usepackage{algorithmic}
\usepackage{algorithm}
\usepackage{array}
\usepackage[caption=false,font=normalsize,labelfont=sf,textfont=sf]{subfig}
\usepackage{textcomp}
\usepackage{stfloats}
\usepackage{url}
\usepackage{verbatim}
\usepackage{graphicx}
\usepackage{cite}
\usepackage{multirow}
\usepackage{tabularx}
\usepackage{booktabs}
\usepackage{ragged2e}
\usepackage{makecell}

\newcolumntype{C}[1]{>{\centering\arraybackslash}m{#1}} 
\newcolumntype{L}[1]{>{\raggedright\arraybackslash}m{#1}} 
\newcolumntype{Y}{>{\centering\arraybackslash}X}

\begin{document}

\title{Towards Next-Generation Healthcare: A Survey of Medical Embodied AI for Perception, Decision-Making, and Action}

\author
{Cheng~Zhang, Qing~Cai,~\IEEEmembership{Member,~IEEE}, Xingzheng Wu, Xun Yang, Xiaojun Chang, Bingkun Bao,~\IEEEmembership{Member,~IEEE}, Liqiang Nie,~\IEEEmembership{Senior Member,~IEEE}, Xinwang Liu,~\IEEEmembership{Senior Member,~IEEE}, Yi Yang,~\IEEEmembership{Fellow,~IEEE}

\IEEEcompsocitemizethanks{\IEEEcompsocthanksitem 

Cheng Zhang and Xingzheng Wu are with the School of Information Science and Engineering, Ocean University of China, Qingdao, Shandong 266100, China (e-mail: zhangcheng@stu.ouc.edu.cn, wuxingzheng@stu.ouc.edu.cn).

Qing Cai is with the Innovation School of Artificial Intelligence, Hefei University of Technology, Hefei 230009, China (e-mail: caiqing1617@gmail.com).

Xun Yang and Xiaojun Chang are with the School of Information Science and Technology, University of Science and Technology of China, Hefei 230026, China (e-mail: xyang21@ustc.edu.cn, xjchang@ustc.edu.cn).

Bing-Kun Bao is with the School of Computer Science and Information Engineering, Hefei University of Technology, Hefei 230009, China (e-mail: bingkunbao@hfut.edu.cn).

Liqiang Nie is with the School of Computer Science and Technology, Harbin Institute of Technology (Shenzhen), Shenzhen 518055, China (e-mail: nieliqiang@gmail.com).

Xinwang Liu is with the College of Computer Science and Technology, National University of Defense Technology, Changsha 410073, China (e-mail: xinwangliu@nudt.edu.cn).

Yi Yang is with the ReLER Laboratory, CCAI, Zhejiang University, Zhejiang 310027, China (e-mail: yangyics@zju.edu.cn).

}
}

\markboth{Journal of \LaTeX\ Class Files,~Vol.~14, No.~8, August~2021}%
{Shell \MakeLowercase{\textit{et al.}}: A Sample Article Using IEEEtran.cls for IEEE Journals}


\maketitle

\begin{abstract}
Foundation models have demonstrated impressive performance in enhancing healthcare efficiency across a wide range of medical applications. Nevertheless, their limited ability to perceive, understand, and interact with the physical world significantly constrains their effectiveness in real-world clinical workflows, where safety-critical decision-making and physical execution are tightly coupled. Recently, embodied artificial intelligence (AI) has emerged as a promising physical-interactive paradigm for intelligent healthcare, enabling agents to operate in complex medical environments. As research in this area rapidly expands, understanding how intelligent agents function as integrated, end-to-end systems in clinical environments becomes increasingly critical. However, existing surveys on medical embodied AI largely emphasize individual aspects or functional components, lacking a unified system-level organization of the field. To support and consolidate recent advances, we systematically survey the core components of medical embodied AI, with a particular emphasis on the coordinated integration of perception, decision-making, and action. We further review representative medical applications and relevant datasets, and we analyze the major challenges encountered in real-world clinical practice. Finally, we discuss key directions for future research in this rapidly evolving field. The associated project can be found at \url{https://github.com/VMVLab/Medical_Embodied_AI_Paper_List}.
\end{abstract}

\begin{IEEEkeywords}
Embodied Artificial Intelligence, Healthcare, Embodied Perception, Embodied Decision-Making, Embodied Action.
\end{IEEEkeywords}

\section{Introduction}

\begin{figure}[t]
\centering
\includegraphics[width=0.5\textwidth]{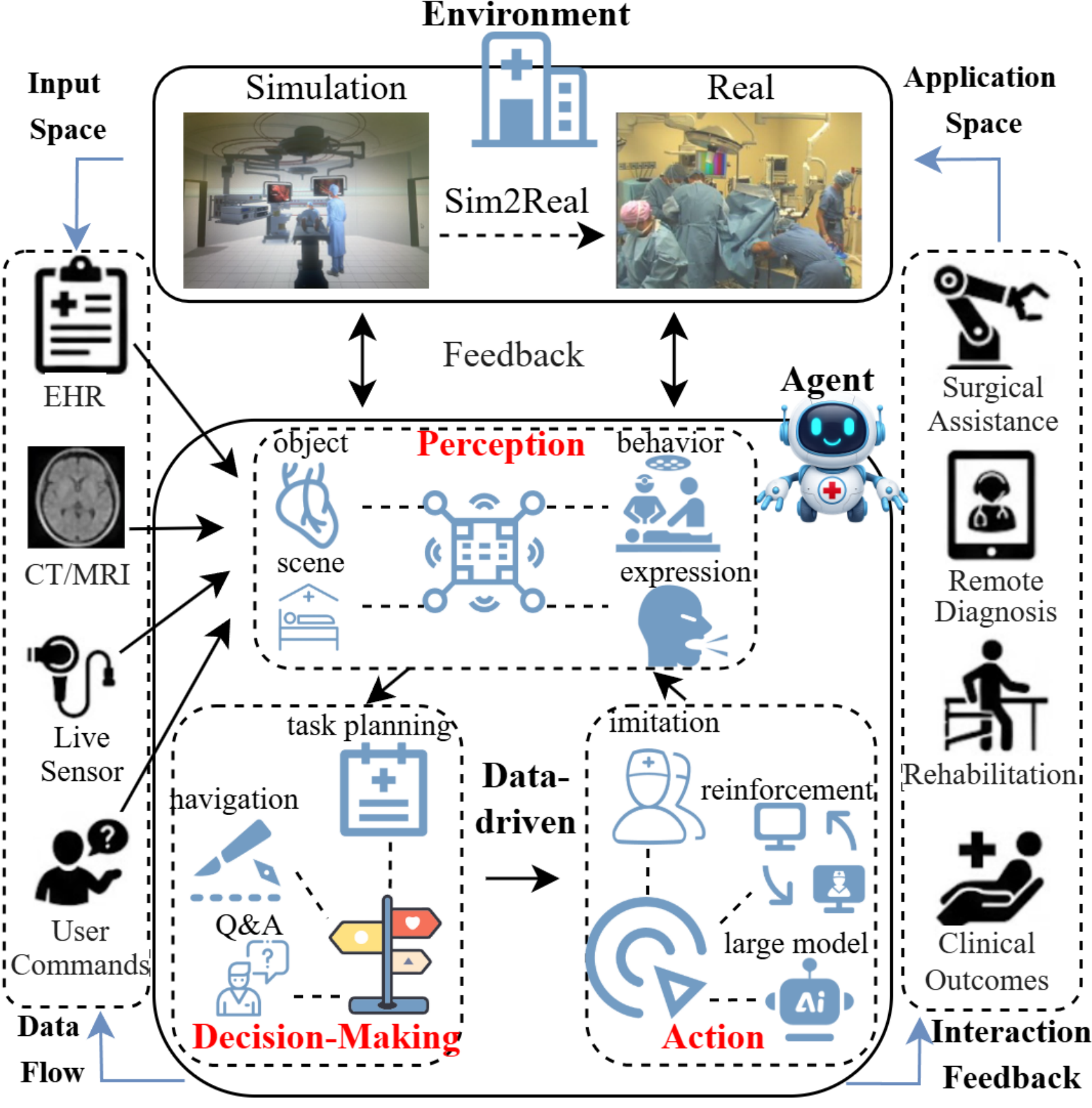}
\caption{Overview of a medical embodied artificial intelligence framework. Medical embodied agents interact with simulated and real clinical environments via a perception–decision–action loop.}
\label{fig1}
\end{figure}

\IEEEPARstart{T}{he} widespread adoption of artificial intelligence (AI) in medicine has significantly improved the efficiency and accuracy of clinical diagnosis \cite{bib2}. Convolutional Neural Networks (CNNs) have achieved strong performance in disease classification and lesion segmentation \cite{bib1,bib3}, while Large Language Models (LLMs) and their multimodal extensions have recently shown promise in medical report generation and clinical decision support \cite{bib4,bib5,bib6,bib7}. However, these approaches are largely confined to a “perception and decision” paradigm based on static data, lacking the capability for physical interaction with real-world clinical environments, which limits their applicability in realistic medical scenarios.

In contrast, embodied artificial intelligence (Embodied AI) enables perception, decision-making, and action within physical environments, opening new avenues for medical AI \cite{bib8,bib9}. As illustrated in Fig.~\ref{fig1}, medical embodied AI systems typically follow a closed-loop perception–decision–action framework. Embodied AI has been applied to a range of medical tasks, including surgical robotics \cite{bib10,bib11}, surgical navigation \cite{bib14,bib15,bib16,bib17,bib18}, rehabilitation assistance \cite{bib20,bib21,bib22}, and mobile clinical support \cite{bib27,bib28,bib29,bib30}, demonstrating clear advantages in complex and dynamic clinical scenarios. Despite this potential, significant challenges remain across embodied perception, decision-making, and action, including data scarcity, uncertainty modeling, and high-precision control sensitivity.

\begin{figure*}[ht]
\centering
\includegraphics[width=1\textwidth]{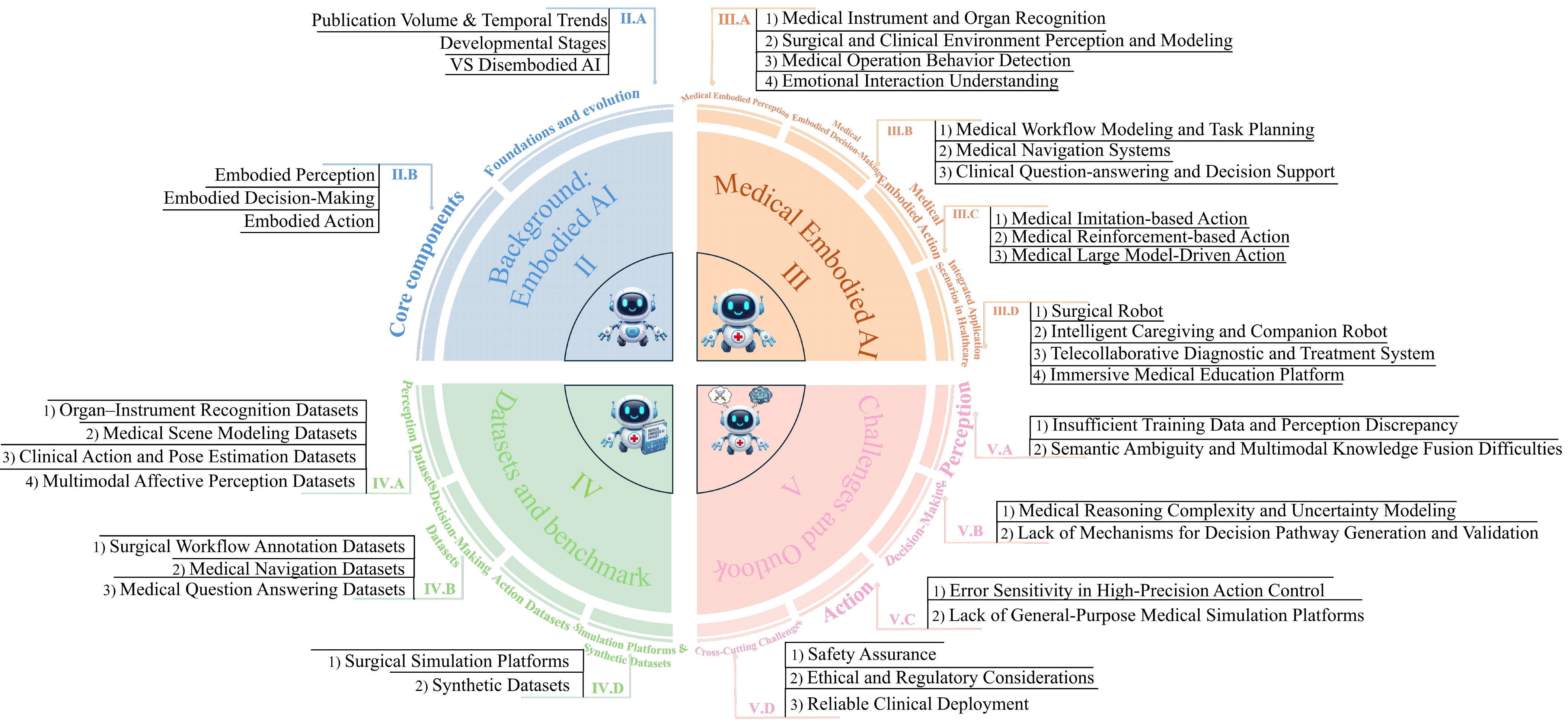}
\caption{Overview structure of this survey.}
\label{fig01}
\end{figure*}

Recent surveys have examined medical embodied AI from complementary perspectives. Some provide broad overviews of embodied AI in healthcare, summarizing functional components, application domains, datasets, and ethical considerations to outline the overall research landscape \cite{bib31}. Others emphasize system-level design, such as hierarchical or modular architectures, to integrate perception, planning, and execution with a focus on clinical reliability and safety \cite{bib34}. Additional works narrow the scope to specific aspects, including representative applications (e.g., surgical robotics and rehabilitation) \cite{bib35}, core perceptual technologies such as 3D medical image segmentation \cite{bib36}, and specialty-driven viewpoints exemplified by ophthalmology \cite{bib37}. Building on these efforts, this work unifies prior perspectives within a closed-loop framework of perception, decision-making, and action, offering a complementary system-level view of medical embodied AI.

\begin{figure*}[ht]
\centering
\includegraphics[width=1\textwidth]{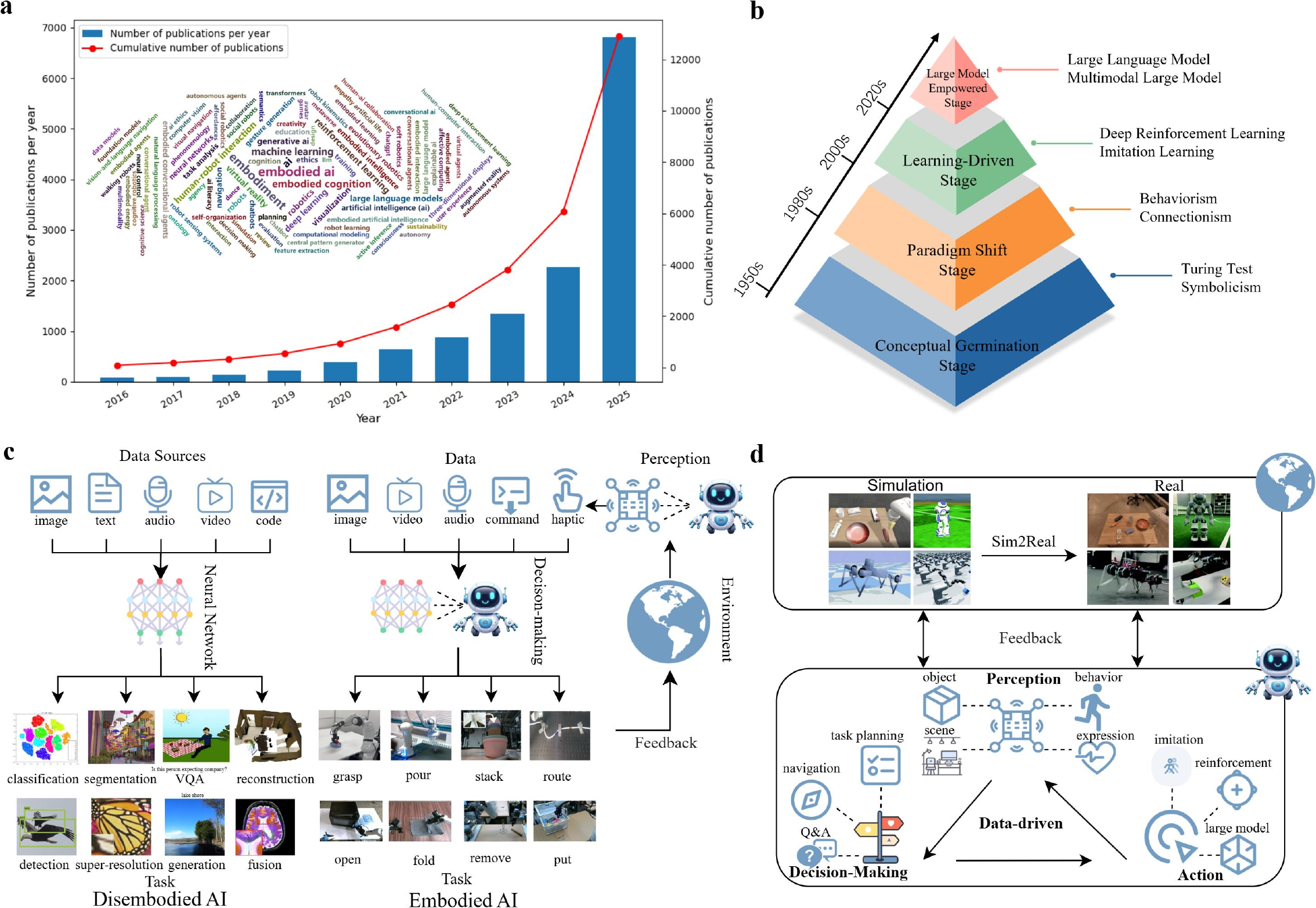}
\caption{Conceptual foundations of embodied AI and its relevance to medical embodied intelligence. a, Publication trends, temporal evolution over the past decade, and representative keywords of embodied AI based on Google Scholar statistics. b, Four developmental stages of embodied AI: conceptual germination, paradigm shift, learning-driven, and large-model–empowered stages. c, Comparison between disembodied intelligence and embodied AI, highlighting the latter’s ability to interact with the environment. d, Core components of embodied AI, including agents and environments at the system level, and embodied perception, decision-making, and action at the technical level.}
\label{fig0_1}
\end{figure*}

As illustrated in Fig.~\ref{fig01}, the remainder of this survey is organized as follows. Section~2 provides background on embodied AI as the conceptual foundation for medical embodied AI, introducing its development and core components. Section~3 examines medical embodied AI applications, while Section~4 introduces relevant datasets. Section~5 discusses key challenges and future perspectives, and Section~6 concludes the survey with key insights and implications for intelligent healthcare systems.

\section{Background: Embodied AI}\label{sec2}

In this section, we briefly review embodied AI as the conceptual foundation of medical embodied AI, focusing on its core ideas, developmental evolution, and system-level components. Rather than providing an exhaustive survey of embodied AI, we aim to establish a concise background that facilitates understanding of subsequent discussions on medical embodied AI.

\subsection{Foundations and Evolution}\label{sec2.1}

Embodied AI has received growing research attention in recent years (Fig.~\ref{fig0_1}a) and has evolved through four major developmental stages (Fig.~\ref{fig0_1}b). The early Conceptual Germination Stage established the foundations of artificial intelligence through symbolic reasoning, followed by a Paradigm Shift Stage that emphasized learning mechanisms and neural networks. The subsequent Learning-Driven Stage leveraged deep reinforcement and imitation learning to enable autonomous decision-making. In the recent Large-Model-Empowered Stage, large language and multimodal models have substantially enhanced perception, cognition, and interaction, exposing the limitations of disembodied AI and motivating embodied systems capable of acting in physical environments.

\begin{table}[htbp]
\centering
\footnotesize
\caption{Overview of the core components, their respective functions, and sub-directions in embodied AI.}
\renewcommand{\arraystretch}{0.9}
\label{tab1}

\begin{tabular}{C{1.9cm} C{2.6cm} C{3.1cm}}
\toprule
\textbf{Components} & \textbf{Function} & \textbf{Sub-Direction} \\
\midrule

\multirow{4}{2cm}{\centering Embodied Perception}
& \multirow{4}{3cm}{Provides multimodal understanding of the environment.}
& Object Perception \\

& & Scene Perception \\

& & Behavior Perception \\

& & Expression Perception \\
\midrule

\multirow{4}{2.2cm}{\centering Embodied Decision-Making}
& \multirow{4}{3.2cm}{Converts perception into adaptive strategies.}
& Task Planning \\

& & Embodied Navigation \\

& & Embodied Question Answering (EQA) \\
\midrule

\multirow{6}{2.2cm}{\centering Embodied Action}
& \multirow{6}{3.2cm}{Executes decisions through physical interaction.}
& Imitation Learning-Based Action \\

& & Reinforcement Learning-Based Action \\

& & Large Model-Driven Action \\
\bottomrule

\end{tabular}
\end{table}

\subsection{Core Components}\label{sec2.2}

Specifically, conventional expert systems and language models operate primarily on abstract or symbolic representations and lack direct interaction with the physical environment, i.e., they are typically disembodied (Fig.~\ref{fig0_1}c). As a result, their adaptability and generalization to complex real-world scenarios remain inherently constrained. In contrast, embodied AI enables agents to perceive, decide, and act in a closed-loop manner with the environment. As illustrated in Fig.~\ref{fig0_1}d, embodied AI systems are typically composed of three core components—embodied perception, embodied decision-making, and embodied action—which jointly support multimodal understanding, planning and reasoning, and autonomous interaction \cite{bib38,bib39,bib40}. In addition, sim-to-real transfer is commonly employed to bridge the gap between simulated training and real-world deployment \cite{bib42,bib43,bib44,bib45}.

Embodied AI typically operates in a closed-loop perception–decision–action paradigm \cite{bib39}. As summarized in Table~\ref{tab1}, embodied perception extracts multimodal representations from heterogeneous sensory inputs (e.g., vision, depth, audio, and touch), supporting object, scene, behavior, and expression understanding for downstream interaction, planning, navigation, and question answering \cite{bib46,bib49}. Building on perceptual representations, embodied decision-making maps observations to adaptive strategies through task planning, navigation, and embodied question answering, enabling goal- and language-aware reasoning \cite{bib99,bib01,bib03,bib02}. Finally, embodied action executes decisions via physical interaction, commonly realized through imitation-based, reinforcement-based, and large-model–driven approaches \cite{bib120}. These properties make embodied AI particularly suitable for safety-critical and environment-dependent medical scenarios.

\begin{figure}[t]
\centering
\includegraphics[width=0.5\textwidth]{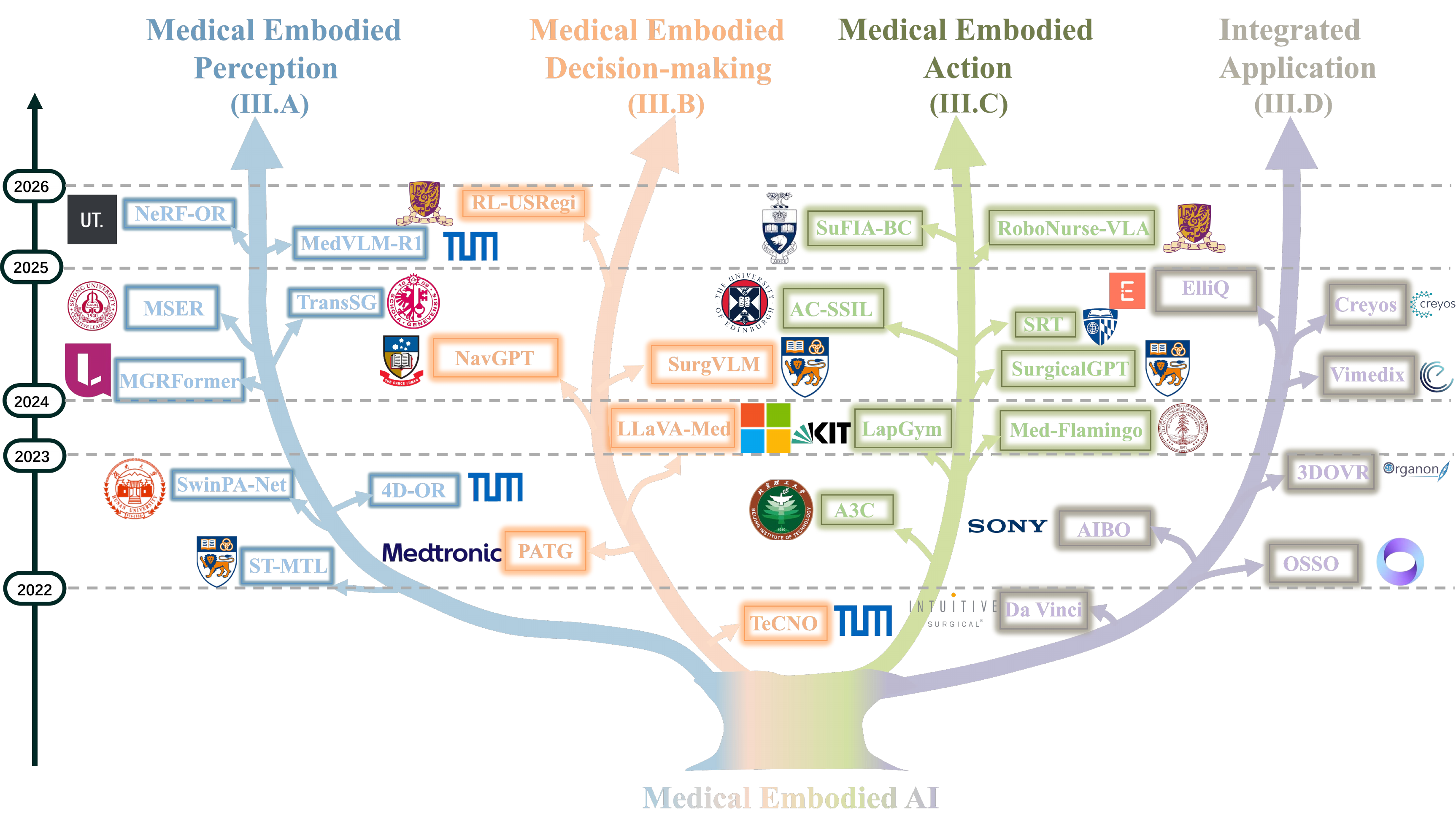}
\caption{Overview of medical embodied AI and its hierarchical organization with representative methods.}
\label{fig2}
\end{figure}

\begin{figure*}[ht]
\centering
\includegraphics[width=1\textwidth]{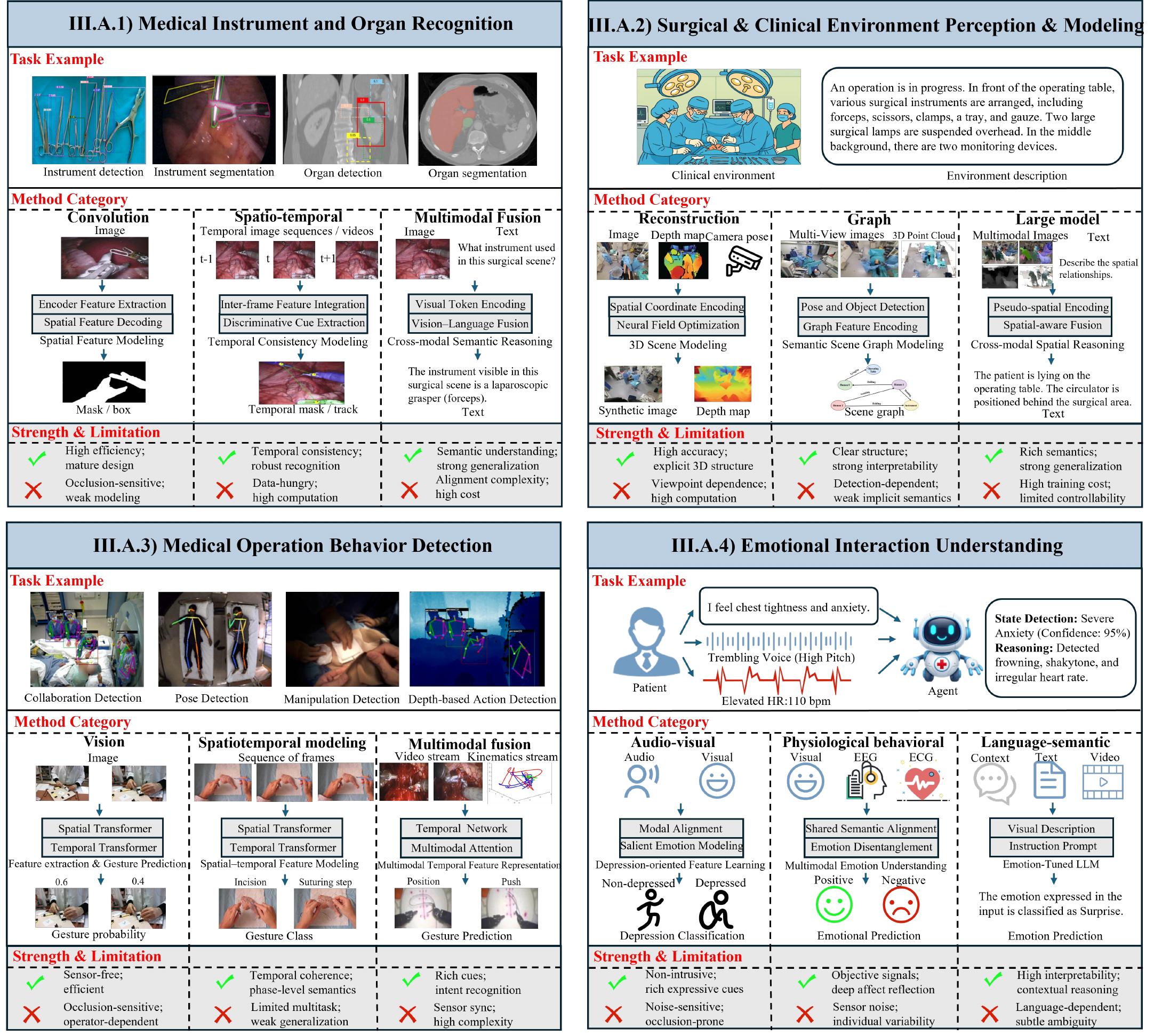}
\caption{Overview of medical embodied perception, including medical instrument and organ recognition, surgical and clinical environment perception and modeling, medical operation behavior detection, and emotional interaction understanding.}
\label{fig02}
\end{figure*}

\section{Medical Embodied AI}\label{sec3}

Building on embodied AI, medical embodied AI has emerged as a paradigm for interactive, task-oriented clinical operation. As illustrated in Fig.~\ref{fig2}, it follows a closed-loop perception–decision–action framework, comprising medical embodied perception, decision-making, and action. Integrated application scenarios instantiate this framework at the system level by jointly combining these components to support real-world medical tasks. 
Accordingly, this chapter reviews representative advances across these aspects to provide an overview of medical embodied AI.

\subsection{Medical Embodied Perception}\label{sec3.1}

Medical embodied perception enables semantic understanding of critical elements in complex medical environments with high object complexity and strict operational constraints. As illustrated in Fig.~\ref{fig02}, this section reviews four key aspects: medical instrument and organ recognition, surgical and clinical environment perception, medical operation behavior detection, and emotional interaction understanding.

\subsubsection{Medical Instrument and Organ Recognition}\label{sec3.1.1}

Medical instrument and organ recognition is a fundamental capability for ensuring operational safety and diagnostic accuracy \cite{bib149,bib150}. Agents must reliably identify diverse surgical tools and complex anatomical structures under challenging conditions, including cluttered scenes, occlusion, blood contamination, unstable illumination, and significant organ deformation, which impose high demands on robustness and real-time performance.

Existing methods can be broadly categorized into three groups, reflecting different strategies for balancing robustness, data dependency, and computational efficiency. Convolution-based image modeling methods are widely used for two- and three-dimensional segmentation of instruments and organs via multi-scale spatial feature modeling. Representative architectures such as U-Net \cite{bib57} and Transformer-based variants (e.g., SwinPA-Net \cite{bib151}) perform well under controlled conditions but, from a robustness perspective, remain sensitive to occlusion, illumination variation, and tissue deformation. Spatio-temporal video modeling methods exploit temporal continuity to capture surgical dynamics and improve stability under motion and transient occlusion; however, compared with convolution-based methods, they typically require large-scale annotated video data and incur higher computational cost \cite{bib152,bib153}. Multimodal fusion–based semantic modeling methods integrate complementary modalities such as vision and language. For example, SurgVLM \cite{bib154} enables prompt-driven recognition of instruments and anatomical structures but, in contrast to purely visual or spatio-temporal methods, faces challenges in cross-modal alignment and inference efficiency.

Discussion: Overall, existing methods address instrument and organ recognition from complementary perspectives, yet their reliance on isolated modeling assumptions limits robustness, efficiency, and generalization under dynamic, safety-critical surgical conditions.

\subsubsection{Surgical and Clinical Environment Perception and Modeling}\label{sec3.1.2}

Surgical and clinical environment perception and modeling aim to provide embodied agents with a structured and global understanding of operating spaces, including room layout, devices, personnel, and dynamic interactions \cite{bib155,bib156}, thereby supporting navigation, collaboration, and task planning.

Existing approaches can be broadly categorized into three groups, reflecting trade-offs among geometric fidelity, semantic abstraction, and computational efficiency. Reconstruction-based methods recover geometric structures from multi-view images, depth data, or point clouds. Approaches such as NeRF-OR \cite{bib157} and Deform3DGS \cite{bib158} achieve high-fidelity reconstruction of surgical environments under static or quasi-static assumptions, but are limited by occlusion, restricted viewpoints, and high computational cost. Recent extensions adopt deformation-aware dynamic 3D Gaussian representations to model non-rigid scenes. Methods such as Endo-HDR \cite{bib1580} and SurgicalGS \cite{bib1581} provide temporally consistent geometry under dynamic motion, improving spatial fidelity for navigation and planning, while still facing challenges in real-time performance, limited supervision, and complex tool–tissue interactions. Graph-based relational methods abstract surgical entities and their interactions into semantic scene graphs. For example, 4D-OR \cite{bib159} models participants, instruments, and spatial relations, while LABRAD-OR \cite{bib160} incorporates temporal memory to capture evolving surgical semantics. These approaches offer more structured and interpretable representations than reconstruction-based methods but depend heavily on accurate entity and relation extraction. Large-model–based semantic understanding methods leverage vision–language or embodied foundation models to infer spatial layouts and semantic relations. For instance, Spatial-ORMLLM \cite{bib161} predicts operating room structure directly from RGB inputs, enabling stronger generalization; however, challenges remain in training cost, cross-modal alignment, and controllability compared with reconstruction- and graph-based approaches.

Discussion: Overall, existing environment perception and modeling methods offer complementary geometric and semantic representations, yet their reliance on isolated reconstruction, relational abstraction, or language-based inference limits robustness and efficiency in dynamic, cluttered surgical environments.

\subsubsection{Medical Operation Behavior Detection}\label{sec3.1.3}

Medical operation behavior detection aims to enable embodied agents to recognize and interpret surgical and clinical actions, providing semantic understanding of procedural workflows, operator intent, and task progression \cite{bib162}. This capability supports real-time feedback, skill assessment, safety monitoring, and decision support in complex medical procedures.

Existing approaches can be broadly categorized into three groups, reflecting different strategies for balancing action granularity, temporal context, and semantic richness. Vision-based action recognition methods identify atomic surgical actions by extracting spatiotemporal features from videos. Representative approaches such as MGRFormer \cite{bib165} and 3D CNN or SlowFast-based models \cite{bib166} effectively capture fine-grained motion patterns but, from a robustness perspective, remain sensitive to occlusion, illumination variation, and operator diversity. Spatiotemporal modeling–based surgical phase inference methods emphasize long-term temporal dependencies to segment procedural workflows. For example, TransSG \cite{bib167} employs spatiotemporal Transformers to recognize gesture sequences, while STANet \cite{bib168} integrates multi-scale temporal features to improve phase recognition; however, compared with vision-based action recognition methods, cross-procedure generalization remains challenging. Multimodal fusion–based semantic behavior understanding methods jointly analyze visual, haptic, auditory, or physiological signals to infer higher-level surgical intent and operator state \cite{bib169,bib170}. In contrast to purely vision-based or spatiotemporal methods, these approaches enhance semantic interpretation but introduce increased sensing complexity and system integration challenges.

Discussion: Overall, existing behavior detection approaches capture complementary aspects of surgical actions, yet their reliance on isolated visual, temporal, or multimodal cues limits robustness and generalization across procedures and operators.

\subsubsection{Emotional Interaction Understanding}\label{sec3.1.4}

Emotional interaction understanding enables embodied agents to perceive affective and intention-related cues in clinical environments, supporting natural and context-aware communication among healthcare staff, patients, and intelligent systems \cite{bib171,bib172}. Such cues are conveyed through speech, facial expressions, body posture, and physiological signals, and are critical for human-centered medical interaction \cite{bib173,bib174}.

Existing approaches can be broadly categorized into three groups, reflecting different strategies for balancing perceptual sensitivity, robustness, and semantic interpretability. Audio-visual–based emotion recognition methods infer emotional states by jointly modeling speech and facial cues. Representative approaches such as DEP-former \cite{bib175} and MSER \cite{bib176} capture dynamic emotional variations in clinical communication but, from a robustness perspective, remain sensitive to noise and occlusion caused by protective equipment. Physiological–behavioral–based affective state estimation methods integrate signals such as heart rate variability, electrodermal activity, and body motion to reflect underlying emotional and stress responses. For example, dual-stream representation learning frameworks \cite{bib177} and multimodal physiological models \cite{bib178} improve recognition accuracy; however, compared with audio-visual methods, they are influenced by sensor noise and individual variability. Language-semantic–based cognitive emotion understanding methods focus on interpreting emotional intent and contextual sentiment in clinical dialogues. Methods such as MedVLM-R1 \cite{bib179} and DialogueLLM \cite{bib180} leverage vision–language or large language models to support emotion-aware reasoning and interaction, but in contrast to perception-driven approaches, their effectiveness depends on high-quality linguistic cues and explicit emotional expression.

Discussion: Overall, existing emotion understanding approaches capture complementary affective cues from audio-visual, physiological, and linguistic signals, yet their reliance on isolated modalities and assumptions limits robustness and consistency in real-world clinical interactions.

\begin{figure*}[ht]
\centering
\includegraphics[width=1\textwidth]{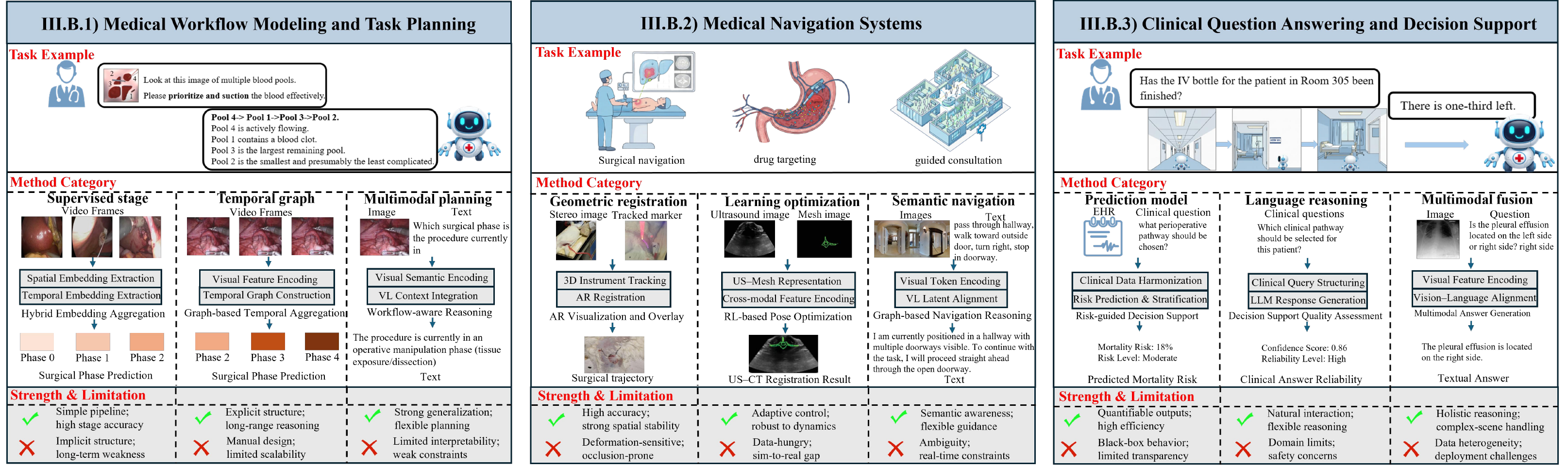}
\caption{Overview of medical embodied decision-making, including medical workflow modeling and task planning, medical navigation systems, and clinical question-answering and decision-support mechanisms.}
\label{fig03}
\end{figure*}

\subsection{Medical Embodied Decision-Making}\label{sec3.2}

Medical embodied decision-making builds on perceptual representations to enable reasoning and planning for clinical tasks \cite{bib181,bib182}, while involving complex temporal dependencies and strong domain constraints. As illustrated in Fig.~\ref{fig03}, this section reviews three directions: medical workflow modeling and task planning, medical navigation, and clinical question answering and decision support.

\subsubsection{Medical Workflow Modeling and Task Planning}\label{sec3.2.1}

Medical workflow modeling and task planning aim to capture procedural structure and task dependencies in surgical or diagnostic processes, enabling agents to infer workflow states and generate high-level action plans \cite{bib183,bib184}. This capability bridges perception and action by supporting temporal reasoning, task decomposition, and policy generation.

Existing approaches can be broadly categorized into three groups, reflecting different strategies for balancing structural explicitness, temporal reasoning, and semantic generalization. Supervised stage-based modeling methods learn workflow segmentation and stage recognition from annotated data using temporal convolution or Transformer architectures. Representative approaches such as Trans-SVNet \cite{bib185} and TeCNO \cite{bib186} achieve stable stage recognition but, from a structural modeling perspective, encode workflow structure implicitly in model parameters, limiting explicit representation of long-term task dependencies. Temporal graph–based task planning methods explicitly model procedural structure by representing stages, tools, or actions as graph nodes with temporal and semantic relations. For example, PATG \cite{bib187} captures cross-stage dependencies via position-aware temporal graphs, while graph-based interaction modeling methods \cite{bib188} encode instrument trajectories over time; however, compared with stage-based methods, they often rely on manually designed graph schemas and task priors. Multimodal semantic planning methods leverage large language or vision–language models to perform high-level task reasoning and decomposition across visual and linguistic modalities. Methods such as SurgVLM \cite{bib154} and LLaVA-Med \cite{bib189} enable language-driven planning with stronger semantic generalization, but in contrast to graph-based approaches, face challenges in interpretability and integrating explicit medical knowledge constraints.

Discussion: Overall, existing workflow modeling and task planning approaches capture complementary aspects of procedural structure and semantic reasoning, yet their reliance on isolated implicit, graph-based, or language-driven representations limits interpretability and robustness in complex clinical workflows.

\subsubsection{Medical Navigation Systems}

Medical navigation systems connect spatial perception with action execution, enabling embodied agents to perform localization, registration, and path planning in diverse clinical scenarios, including surgical robotics, interventional procedures, and in-hospital guidance.

\paragraph{Surgical Robotics and Intraoperative Navigation}
In surgical settings, navigation systems emphasize precise localization and registration among patients, instruments, and preoperative images, with geometric and image-registration–based methods remaining dominant. Systems such as the BrainLab VectorVision Neuronavigation System \cite{bib194} and augmented-reality–based platforms \cite{bib195} achieve high accuracy but remain sensitive to tissue deformation, occlusion, and real-time constraints.

\paragraph{Interventional Navigation}
For minimally invasive and image-guided interventions, navigation systems must adapt to dynamic anatomy and limited sensing conditions. Learning- and optimization-based approaches introduce machine learning and reinforcement learning to improve adaptability. For example, RL-USRegi \cite{bib196} enables autonomous ultrasound registration, while inverse reinforcement learning–based methods support catheter and guidewire navigation by imitating expert behavior \cite{bib197}. These approaches enhance flexibility but typically incur high training cost and face sim-to-real transfer challenges.

\paragraph{In-hospital Guidance and Semantic Navigation}
Beyond procedural navigation, embodied agents are increasingly used for in-hospital guidance. Multimodal semantic methods integrate visual, linguistic, and spatial cues to support language-driven, environment-aware navigation. Systems such as SurgVLM \cite{bib154}, NavGPT \cite{bib198}, and NavGPT-2 \cite{bib199} demonstrate this capability but face challenges in semantic ambiguity, cross-modal alignment, and real-time performance. Recent studies combine bird’s-eye-view perception or scene maps with large language models to improve instruction generation and controllability \cite{bib04,bib05}.

Discussion: Overall, existing medical navigation systems address complementary aspects of localization, planning, and semantic guidance, yet their reliance on isolated geometric, learning-based, or language-driven paradigms limits robustness and real-time reliability in dynamic clinical environments.

\begin{figure*}[ht]
\centering
\includegraphics[width=1\textwidth]{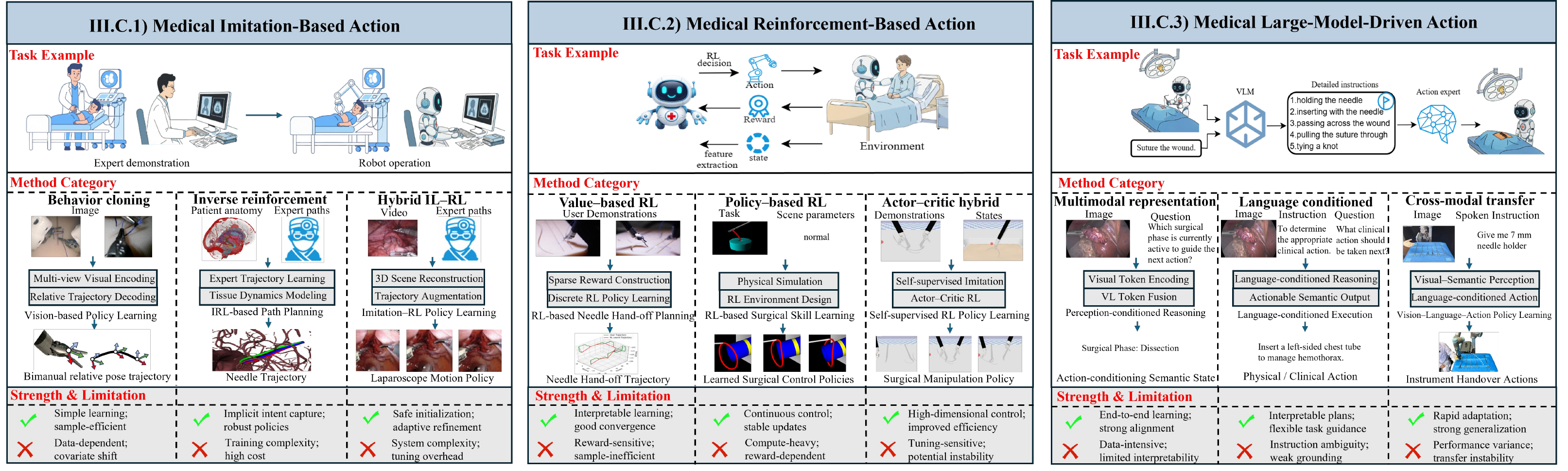}
\caption{Overview of medical embodied action, including medical imitation-based action, medical reinforcement-based action, and medical large-model-driven action.}
\label{fig04}
\end{figure*}

\subsubsection{Clinical Question Answering and Decision Support}

Clinical question answering and decision support enable embodied agents to reason over multimodal clinical information and provide interpretable recommendations or action guidance, bridging task planning and execution in real-world medical settings \cite{bib201,bib202}.

Existing approaches can be broadly categorized into three groups, reflecting different strategies for balancing predictive accuracy, interpretability, and interaction flexibility. Prediction-based decision support methods apply machine learning models to clinical records or imaging data for risk assessment, outcome prediction, and treatment planning. While providing quantifiable decision cues, they often lack transparency and interactive reasoning capability \cite{bib203}. Language-model–based decision support methods leverage natural language question answering to assist case interpretation and clinical decision-making, improving human–machine communication; however, compared with prediction-based methods, they remain constrained by medical specialization, interpretability, and safety concerns \cite{bib204}. Multimodal fusion–based decision support methods integrate imaging, clinical text, behavioral, and physiological signals to enable comprehensive reasoning across patient states and clinical workflows \cite{bib205,bib206}. In contrast to unimodal prediction or language-based approaches, these methods enhance contextual completeness but face challenges in data standardization, real-time processing, and deployment reliability.

Discussion: Overall, existing clinical question answering and decision support approaches provide complementary predictive and reasoning capabilities, yet their reliance on isolated predictive, language-based, or multimodal frameworks limits interpretability, safety, and reliable deployment in real-world clinical settings.

\subsection{Medical Embodied Action}\label{sec3.3}

Medical embodied action focuses on executing perceptual and decision outputs through physical interaction, enabling embodied agents to autonomously perform medical procedures under strict precision and safety constraints \cite{bib207,bib208}. As illustrated in Fig.~\ref{fig04}, this section reviews three representative paradigms: medical imitation-based action, medical reinforcement-based action, and medical large-model-driven action.

\subsubsection{Medical Imitation-Based Action}\label{sec3.3.1}

Medical imitation-based action aims to transfer expert surgical skills to embodied agents through demonstrations, providing a safe and sample-efficient alternative to trial-and-error learning in high-risk clinical settings \cite{bib209}.

Existing approaches can be broadly categorized into three groups, reflecting trade-offs among sample efficiency, robustness, and adaptability. Behavior cloning (BC) learns state–action mappings from expert demonstrations and is widely used for basic surgical skills. Methods such as the Surgical Robot Transformer (SRT) \cite{bib210} and Intermittent Visual Servoing \cite{bib211} achieve effective imitation via relative-action representations and visual closed-loop control, while SuFIA-BC \cite{bib212} improves generalization through synthetic demonstrations; however, BC remains sensitive to demonstration quality and distributional shift. Inverse reinforcement learning (IRL) and adversarial imitation learning infer implicit expert objectives to produce more robust policies. Examples include steerable-needle path planning with learned reward functions \cite{bib213} and adversarial trajectory imitation for catheter insertion \cite{bib214}. Compared with BC, these methods improve robustness but incur higher training complexity and computational cost. Hybrid imitation–reinforcement learning combines demonstration-based initialization with reinforcement refinement to enhance adaptability. Systems such as surgeon-preference-aware ophthalmic assistants \cite{bib215} and the ILLC framework for laparoscope control \cite{bib216} enable personalized adaptation and cross-scenario generalization, but introduce additional system complexity and tuning overhead.

Discussion: Overall, imitation-based action approaches capture complementary strengths in sample efficiency, robustness, and adaptability, yet their reliance on fixed demonstrations or implicit reward assumptions limits generalization and stability under distributional shift in real-world clinical scenarios.

\subsubsection{Medical Reinforcement-Based Action}\label{sec3.3.2}

Medical reinforcement-based action leverages reinforcement learning to optimize control policies through interaction with medical environments, enabling embodied agents to achieve adaptive and autonomous execution beyond imitation-based strategies \cite{bib217}. This paradigm is particularly suited for complex surgical tasks requiring continuous control and dynamic adaptation.

Existing approaches can be broadly categorized into three groups, reflecting different strategies for balancing sample efficiency, control stability, and scalability. Value-based methods learn state–action value functions to guide policy optimization and are suitable for relatively low-dimensional decision problems. Representative work such as Collaborative Suturing \cite{bib218} applies Q-learning to enable autonomous handover actions but, from a reward design perspective, remains sensitive to handcrafted rewards and limited sample efficiency. Policy-based methods directly optimize control policies for continuous action spaces and exhibit stable learning behavior. For example, LapGym \cite{bib219} employs PPO to learn laparoscopic manipulation policies, while A3C-based approaches support navigation and instrument control in virtual intervention settings \cite{bib220}; however, compared with value-based methods, they typically require substantial training data and computational resources. Actor–Critic methods combine value estimation with policy learning to improve stability and efficiency in high-dimensional environments. Representative approaches such as AC-SSIL \cite{bib221} and CASOG \cite{bib225} integrate imitation guidance or conservative value estimation to enhance robustness and sample efficiency, but in contrast to purely policy-based approaches, demand careful tuning to balance dual learning objectives.

Discussion: Overall, reinforcement-based action methods provide strong adaptability and control flexibility, yet their reliance on carefully designed rewards and large-scale interaction limits sample efficiency, stability, and practical deployment in real-world clinical environments.

\subsubsection{Medical Large-Model-Driven Action}\label{sec3.3.3}

Medical large-model-driven action leverages vision–language–action foundation models to map high-level semantic understanding to executable medical actions, enabling embodied agents to perform complex tasks with strong generalization under limited supervision \cite{bib222,bib223,bib224}.

Existing approaches can be broadly categorized into three paradigms, reflecting different strategies for balancing alignment fidelity, planning flexibility, and adaptation efficiency. Multimodal perception–action alignment methods jointly model visual observations and action trajectories to enable end-to-end action generation. Representative models such as SurgicalGPT \cite{bib226} and SurgVLM \cite{bib154} integrate surgical videos and language supervision to predict action sequences and procedural steps but, from a controllability perspective, remain sensitive to cross-modal misalignment and execution uncertainty. Language-conditioned task planning methods utilize natural language instructions or structured knowledge to guide action generation under semantic constraints. For example, LLaVA-Med \cite{bib189} and Med-Flamingo \cite{bib227} enable instruction-driven execution for tasks such as instrument alignment and path planning; however, compared with perception–action alignment methods, they rely more heavily on accurate language grounding and task specification. Cross-modal transfer–based few-shot execution methods focus on rapid adaptation to unseen tasks through parameter-efficient tuning or sim-to-real transfer. Systems such as RoboNurse-VLA \cite{bib228} demonstrate effective real-time instrument manipulation under verbal guidance, but in contrast to planning-driven approaches, often trade explicit task reasoning for adaptability and scalability.

Discussion: Overall, large-model-driven action methods offer strong semantic generalization and planning flexibility, yet their reliance on cross-modal alignment and implicit reasoning limits controllability, reliability, and safe execution in real-world clinical settings.

\begin{table*}[t]
\caption{Integrated application systems of medical embodied AI.}
\label{tab5}
\centering
\fontsize{7.2}{8.4}\selectfont
\setlength{\tabcolsep}{3pt}
\renewcommand{\arraystretch}{1.0}

\begin{tabularx}{\textwidth}{@{} 
C{2cm} 
L{3cm} 
L{1.4cm} 
Y 
C{1.8cm} 
@{}}
\toprule
\textbf{Category} & 
\textbf{System/Robot} & 
\textbf{Setting} & 
\textbf{Core functions} & 
\textbf{Autonomy} \\
\midrule

\multirow{10}{*}{\parbox{1.6cm}{\centering Surgical\\Robot}} &
da Vinci~\cite{bib229} & OR & Surgeon-controlled manipulation; improves precision and consistency & Teleop \\
& EMARO~\cite{bib232} & OR & Endoscope stabilization and positioning; reduces assistant burden & Shared \\
& ROSA~\cite{bib233} & OR & Image-guided navigation and positioning; improves placement accuracy & Assisted \\
& PRECEYES~\cite{bib234} & OR & Micron-level manipulation; tremor suppression & Teleop \\
& ROBODOC~\cite{bib235} & OR & High-precision bone preparation for improved implant fit & Supervised \\
& Symani~\cite{bib236} & OR & Enhanced micro-scale dexterity and consistency & Teleop \\
& CyberKnife~\cite{bib237} & RT suite & Image-guided adaptive radiation delivery & Automated \\
& Monarch~\cite{bib238} & Endoscopy & Stable access for peripheral lung sampling & Assisted \\
& Flex~\cite{bib239} & OR & Flexible access in confined anatomy & Teleop \\
& CorPath GRX~\cite{bib240} & Cath lab & Precise device manipulation; reduced radiation exposure & Teleop \\
\midrule

\multirow{8}{*}{\parbox{1.6cm}{\centering Intelligent\\Caregiving\\\& Companion\\Robot}} &
PARO~\cite{bib241} & Ward/Home & Emotional interaction and engagement & Interactive \\
& AIBO~\cite{bib242} & Home & Social companionship and engagement & Interactive \\
& Pepper~\cite{bib243} & Hospital & Social interaction and basic assistance & Interactive \\
& ElliQ~\cite{bib244} & Home & Daily support, reminders, and engagement & Proactive \\
& Arash~\cite{bib245} & Hospital & Affective interaction; anxiety reduction & Interactive \\
& Robear~\cite{bib246} & Care & Physical assistance (e.g., lifting, transfer) & Assisted \\
& Giraff~\cite{bib247} & Telecare & Remote presence for distributed care & Telepresence \\
& Telenoid~\cite{bib248} & Telecare & Embodied communication proxy & Telepresence \\
\midrule

\multirow{6}{*}{\parbox{1.6cm}{\centering Immersive\\Medical\\Education\\Platform}} &
Touch Surgery~\cite{bib249} & Training & Procedural cognition and decision rehearsal & -- \\
& Body Interact~\cite{bib250} & Training & Decision-making training with feedback & -- \\
& VirtaMed ArthroS~\cite{bib251} & Training & Operative skill learning and assessment & -- \\
& Vimedix 3.2~\cite{bib252} & Training & Scanning skill training and evaluation & -- \\
& OSSO~\cite{bib253} & Training & Skill acquisition and competency tracking & -- \\
& 3D Organon VR~\cite{bib254} & Training & Spatial anatomy understanding & -- \\
\midrule

\multirow{4}{*}{\parbox{1.6cm}{\centering \tiny Telecollabo-rative Diagnostic Treatment System}} &
Teladoc Mini Cart~\cite{bib255} & Remote & Remote examination and consultation & Telepresence \\
& Mercy Telehealth~\cite{bib256} & Tele-ICU & Distributed monitoring and decision support & Telemedicine \\
& VSee~\cite{bib257} & Telehealth & Multimodal communication and care coordination & Telemedicine \\
& Creyos~\cite{bib258} & Remote & Cognitive assessment and longitudinal monitoring & Workflow \\
\bottomrule
\end{tabularx}
\end{table*}

\subsection{Integrated Application Scenarios in Healthcare}\label{sec3.4}

Integrated application scenarios demonstrate how medical embodied AI realizes closed-loop perception, decision-making, and action in real clinical settings, marking a key step toward practical healthcare deployment. This section reviews four application domains: surgical robots, intelligent caregiving and companion robots, immersive medical education platforms, and telecollaborative diagnostic and treatment systems (Table~\ref{tab5}).

\subsubsection{Surgical Robot}\label{subsubsec3}

Surgical robots represent the most mature and widely deployed form of medical embodied AI, integrating multimodal perception, intelligent decision-making, and precise action execution to support complex clinical procedures. These systems are extensively applied in minimally invasive surgery, neurosurgery, orthopedics, ophthalmology, and interventional medicine, where high precision and safety are required.

Representative platforms include the da Vinci Surgical System~\cite{bib229,bib230,bib231}, which dominates minimally invasive surgery through teleoperated multi-degree-of-freedom manipulation and high-definition 3D vision; EMARO~\cite{bib232} and ROSA~\cite{bib233} provide image-guided assistance and autonomous positioning in endoscopic and neurosurgical procedures; PRECEYES~\cite{bib234} and Symani~\cite{bib236} enable microscale manipulation via motion scaling and tremor suppression; and CyberKnife~\cite{bib237} integrates real-time imaging with robotic radiosurgery. Flexible and catheter-based systems, such as Monarch~\cite{bib238}, Flex~\cite{bib239}, and CorPath GRX~\cite{bib240}, further extend embodied intelligence to pulmonary, transoral, and cardiovascular interventions. Together, these systems illustrate how embodied AI principles are instantiated in real-world surgical practice.

\begin{figure}[ht]
\centering
\includegraphics[width=0.5\textwidth]{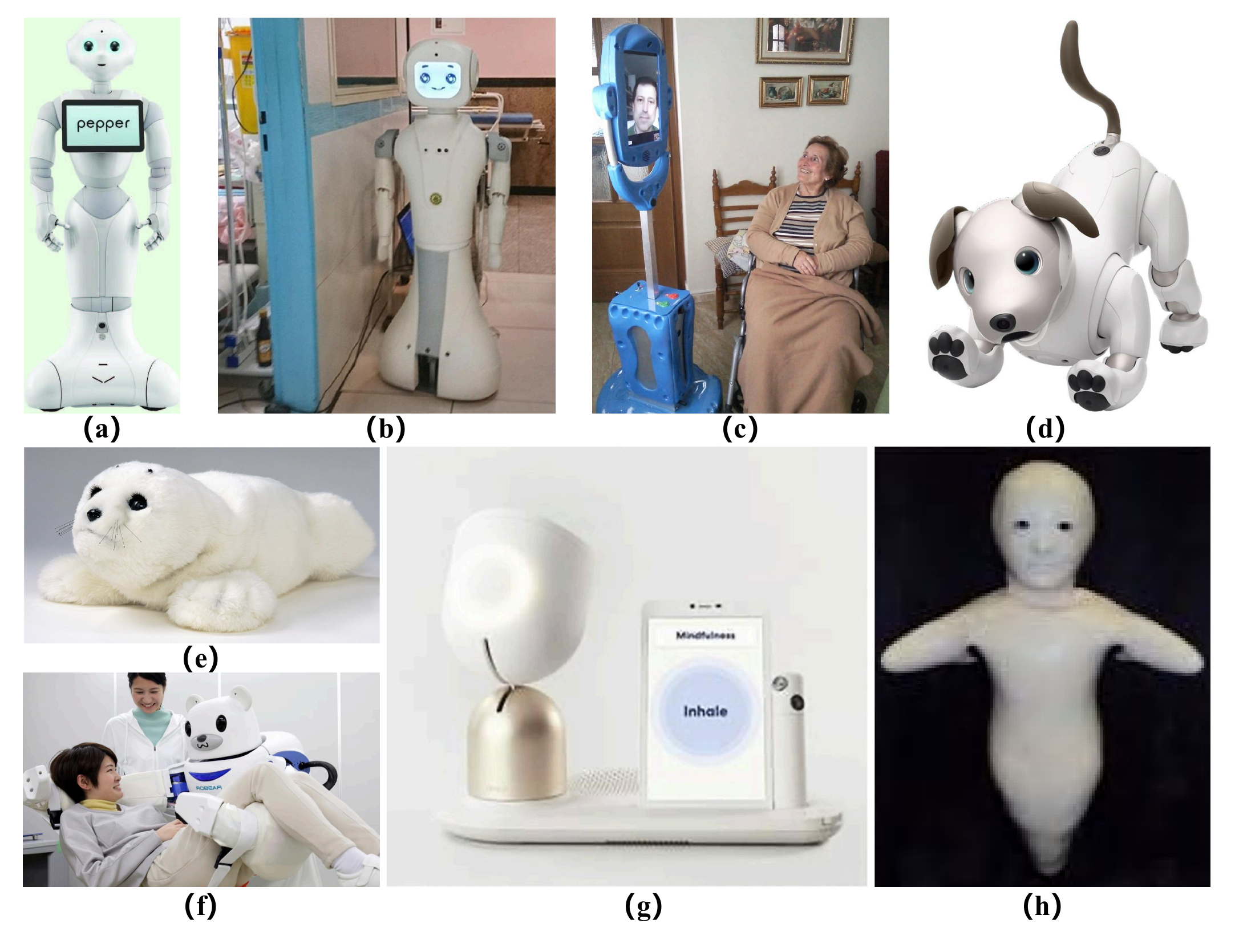}
\caption{Intelligent Caregiving and Companion Robot. (a) Pepper \cite{bib243}. (b) Arash \cite{bib245}.(c) Giraff \cite{bib247}. (d) AIBO \cite{bib242}. (e) PARO \cite{bib241}. (f) Robear \cite{bib246}. (g) ElliQ \cite{bib244}. (h) Telenoid \cite{bib248}.}
\label{fig4}
\end{figure}

\subsubsection{Intelligent Caregiving and Companion Robot}\label{sec3.4.1}

Intelligent caregiving and companion robots extend medical embodied AI from clinical procedures to daily care and long-term assistance, providing continuous support in hospital wards, rehabilitation centers, and eldercare environments. As shown in Fig.\ref{fig4}, these systems emphasize robust human–robot interaction, contextual awareness, and emotional engagement while complying with medical safety and ethical constraints.

Existing systems can be broadly categorized into three groups. Emotional companion robots focus on social interaction and affective support through multimodal perception and dialogue. Representative platforms such as PARO \cite{bib241}, AIBO \cite{bib242}, Pepper \cite{bib243}, ElliQ \cite{bib244}, and Arash \cite{bib245} have been deployed in dementia care, pediatric wards, and home-based eldercare to enhance emotional well-being and engagement. Physical assistance robots provide direct support for patient transfer, posture adjustment, and mobility assistance using compliant control and safe human–robot interaction mechanisms. Robear \cite{bib246} exemplifies this category by enabling stable lifting and transfer of patients with limited mobility. Telepresence companion robots facilitate remote caregiving and social connection by integrating communication interfaces with embodied mobility. Systems such as Giraff \cite{bib247} and Telenoid \cite{bib248} support remote ward rounds, monitoring, and emotional communication, extending the reach of caregivers and clinicians.

\subsubsection{Immersive Medical Education Platform}\label{sec3.4.2}

Immersive medical education platforms leverage virtual, augmented, and mixed reality technologies to provide safe, repeatable, and standardized training for medical education, overcoming the limitations of traditional experience-based instruction. By reconstructing anatomical structures, clinical procedures, and pathological processes, these platforms support efficient skill acquisition, remote learning, and objective assessment.

Existing systems can be broadly categorized into three groups. Cognitive training platforms emphasize clinical reasoning and decision-making through interactive simulations; representative systems such as Touch Surgery \cite{bib249} and Body Interact \cite{bib250} support rehearsal of procedural logic, case analysis, and emergency decision-making in virtual scenarios. Operative skill training platforms focus on hands-on procedural practice with realistic visual and haptic feedback. Platforms including VirtaMed ArthroS \cite{bib251} and Vimedix 3.2 \cite{bib252} enable arthroscopy and ultrasound training via high-fidelity simulation and automated performance evaluation. Comprehensive simulation platforms integrate anatomy visualization with multidisciplinary education. Systems such as OSSO \cite{bib253} and 3D Organon VR Anatomy \cite{bib254} provide immersive exploration of anatomical structures and spatial relationships, supporting standardized education and foundational skill development across medical disciplines.

\subsubsection{Telecollaborative Diagnostic and Treatment System}\label{sec3.4.3}

Telecollaborative diagnostic and treatment systems enable distributed clinical collaboration by integrating sensing, communication, and interaction technologies, thereby overcoming geographic constraints and improving access to high-quality healthcare services. These systems support remote consultation, monitoring, and decision-making in scenarios such as emergency response, primary care assistance, and multidisciplinary collaboration.

Representative platforms include Teladoc Mini Cart \cite{bib255} facilitates real-time clinician–patient interaction for telemedicine services; Mercy Telehealth \cite{bib256}, which enables continuous remote monitoring and guidance in intensive care settings; and VSee \cite{bib257} supports secure, low-bandwidth video-based collaboration for pathology consultation and medical education. In addition, Creyos \cite{bib258} provides remote cognitive assessment and quantitative analysis, illustrating the extension of telecollaborative systems to specialized diagnostic tasks. Together, these systems demonstrate how embodied AI technologies can be integrated into distributed healthcare workflows to enhance collaboration efficiency and resource sharing.

\begin{figure*}[ht]
\centering
\includegraphics[width=1\textwidth]{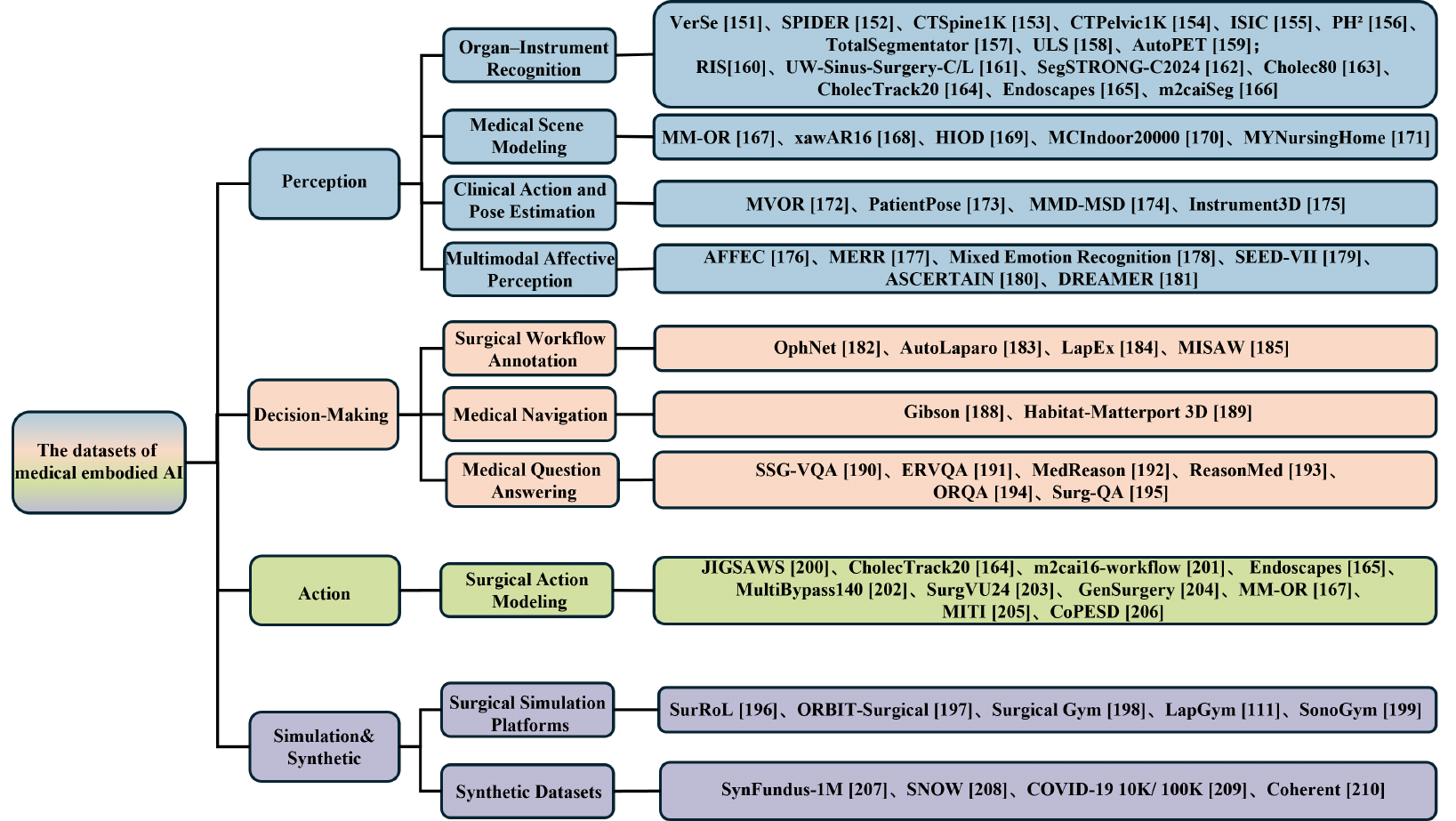}
\caption{The datasets of medical embodied AI, which are categorized into perception, decision-making, action, and simulation \& synthetic.}
\label{fig3}
\end{figure*}

\section{Datasets and Benchmarks}\label{sec4}

High-quality datasets are essential for advancing medical embodied AI. This section reviews representative publicly available datasets, organized according to the three technical layers of perception, decision-making, and action, together with simulation and synthetic data resources (Fig.~\ref{fig3}).

\subsection{Perception Datasets}\label{sec4.1}

\subsubsection{Organ and Instrument Recognition Datasets}\label{sec4.1.1}

Organ recognition datasets primarily support segmentation and classification across diverse anatomical regions. Public benchmarks cover major organs and pathological structures in the brain, thorax, abdomen, musculoskeletal system, and skin. Representative datasets include VerSe, SPIDER, CTSpine1K, and CTPelvic1K for spinal and skeletal analysis \cite{bib268,bib269,bib270,bib271}, ISIC and PH$^2$ for dermatological lesion analysis \cite{bib272,bib274}, and TotalSegmentator \cite{bib277} for large-scale multi-organ annotation. For whole-body and oncological analysis, datasets such as ULS \cite{bib278} and AutoPET \cite{bib279} support automated tumor segmentation and systemic disease assessment.

Medical instrument recognition datasets focus on detecting, segmenting, and tracking surgical tools, forming a foundation for surgical automation and intraoperative assistance. Dedicated datasets include RIS \cite{bib280} for robotic instruments, UW-Sinus-Surgery-C/L \cite{bib281} for endoscopic sinus surgery, and SegSTRONG-C2024 \cite{bib282}, which emphasizes robustness under challenging imaging conditions. In addition, large-scale surgical video datasets such as Cholec80 \cite{bib283}, CholecTrack20 \cite{bib284}, Endoscapes \cite{bib285}, and m2caiSeg \cite{bib286} provide multi-task annotations for instrument recognition, tracking, and scene understanding.

\subsubsection{Medical Scene Modeling Datasets}\label{sec4.1.2}

Medical scene modeling datasets support the perception and understanding of spatial layouts, equipment distribution, and human activities in clinical environments such as operating rooms and hospital wards. These datasets enable research ranging from semantic perception to 3D reconstruction and environment-aware interaction.

Representative datasets include MM-OR \cite{bib288}, a large-scale multimodal benchmark for operating room scene understanding that supports tasks such as semantic segmentation and scene graph construction. The xawAR16 dataset \cite{bib289} provides RGB-D images and precise pose annotations for evaluating visual localization and 3D mapping in mixed-reality surgical environments. HIOD \cite{bib290} and MCIndoor20000 \cite{bib291} focus on object detection and structural recognition in hospital interiors, supporting indoor perception and navigation research. In addition, MYNursingHome \cite{bib292} targets elderly care scenarios, enabling scene understanding and assistive interaction in long-term care environments.

\subsubsection{Clinical Action and Pose Estimation Datasets}\label{sec4.1.3}

Clinical action and pose estimation datasets support the modeling of human motion, posture, and interactive behaviors in medical environments, enabling embodied agents to understand clinical activities during diagnosis, nursing, and surgical procedures.

Representative datasets include MVOR \cite{bib293}, which provides multi-view RGB-D recordings from real operating rooms for 3D pose estimation and multi-person tracking; PatientPose \cite{bib294} offers upper-body pose annotations from long-term clinical recordings to support patient motion analysis; and MMD-MSD \cite{bib295}, which integrates vision and wearable sensor data to model posture and physiological states in healthcare-related activities. In addition, Instrument3D \cite{bib296} supports precise 3D tracking of surgical instruments, facilitating fine-grained analysis of surgical actions.

\subsubsection{Multimodal Affective Perception Datasets}\label{sec4.1.4}

Multimodal affective perception datasets provide synchronized signals from physiological, behavioral, and visual modalities to support emotion recognition, affective modeling, and human–computer interaction in embodied AI.

Representative datasets include AFFEC \cite{bib297}, which integrates eye tracking, facial action units, galvanic skin response, and personality traits for multimodal emotion classification; MERR \cite{bib298} and Mixed Emotion Recognition \cite{bib299}, which support coarse- and fine-grained emotion recognition from multimodal signals; and SEED-VII \cite{bib300} offers EEG and eye-tracking data with continuous emotion intensity labels. In addition, ASCERTAIN \cite{bib301} and DREAMER \cite{bib302} combine EEG, ECG, and peripheral physiological signals with affective annotations, enabling research on emotion–personality relationships and practical emotion recognition in real-world settings.

\subsection{Decision-Making Datasets}\label{sec4.2}

\subsubsection{Surgical Workflow Annotation Datasets}\label{sec4.2.1}

Surgical workflow annotation datasets provide structured temporal labels for modeling procedural phases, task dependencies, and decision logic, forming a key data foundation for surgical automation and decision-making.

Representative datasets include the CHOLECT dataset series \cite{bib284}, which supports fine-grained action recognition through triplet annotations of instruments, actions, and targets in laparoscopic cholecystectomy; OphNet \cite{bib303} offers hierarchical phase and action annotations for ophthalmic surgeries; and AutoLaparo \cite{bib304}, which integrates workflow recognition with motion prediction and image segmentation for hysterectomy procedures. In addition, LapEx \cite{bib305} focuses on sleeve gastrectomy with activity, scene, and skill assessment labels, while MISAW \cite{bib306} combines synchronized video and kinematic data with phase-level annotations to support multimodal analysis of minimally invasive vascular anastomosis.

\subsubsection{Medical Navigation Datasets}\label{sec4.2.2}

Medical navigation datasets support localization, path planning, and spatial understanding for embodied agents in surgical and clinical environments.

Representative datasets include the Portable 6D Surgical Instrument Magnetic Localization Dataset \cite{bib307}, which provides six-degree-of-freedom instrument tracking data for minimally invasive surgical navigation; the Head Model Collection for Mixed Reality Navigation \cite{bib308}, which offers CT/MRI-derived anatomical models for mixed-reality neurosurgical guidance; and large-scale 3D environment datasets such as Gibson \cite{bib309} and Habitat-Matterport 3D \cite{bib310}, which provide realistic indoor digital twins to support research on navigation, spatial reasoning, and autonomous planning in medical and hospital-like settings.

\subsubsection{Medical Question Answering Datasets}\label{sec4.2.3}

Medical question answering datasets provide high-level semantic understanding and reasoning supervision for embodied AI, supporting clinical decision-making and multimodal interaction.

Representative datasets include SSG-VQA \cite{bib311}, which constructs visual question answering benchmarks from laparoscopic videos using structured surgical scene graphs; and ERVQA \cite{bib312}, which focuses on emergency room scenarios to evaluate vision–language reasoning in real clinical environments. Knowledge-centric reasoning datasets such as MedReason \cite{bib313} and ReasonMed \cite{bib314} emphasize stepwise logical inference and interpretable medical reasoning based on structured knowledge and large language models. In addition, ORQA \cite{bib315} integrates multimodal data from operating room environments to support multitask surgical question answering, while Surg-QA \cite{bib316} provides large-scale instruction-based video question answering to enable semantic understanding of complex surgical workflows.

\begin{table*}[t]
\centering
\caption{Representative simulation platforms for medical embodied AI.}
\label{tab6}
\scriptsize
\renewcommand{\arraystretch}{1.25}
\setlength{\tabcolsep}{5pt}

\newcolumntype{L}[1]{>{\RaggedRight\arraybackslash}m{#1}}
\newcolumntype{C}[1]{>{\centering\arraybackslash}m{#1}}

\begin{tabularx}{\textwidth}{@{} C{3cm} C{3.5cm} C{3.3cm} C{3.5cm} C{3.0cm} @{}}
\toprule
\textbf{Platform} &
\textbf{Representative Tasks} &
\textbf{Learning Paradigms} &
\textbf{Strength} &
\textbf{Limitation} \\
\midrule

\makecell[c]{SurRoL \cite{bib326}} &
Grasping, cutting, suturing, general robotic surgery &
\makecell[c]{Imitation learning;\\ reinforcement learning} &
Strong physical realism; diverse surgical tasks &
Moderate visual fidelity; high compute cost \\
\midrule

\makecell[c]{ORBIT-Surgical \cite{bib327}} &
Surgical dexterity learning; active perception &
\makecell[c]{Reinforcement learning;\\ imitation learning} &
Excellent visual realism; efficient GPU-parallel simulation &
High hardware demand; complex system setup \\
\midrule

\makecell[c]{Surgical Gym \cite{bib328}} &
Massive RL training; rapid policy iteration &
Reinforcement learning &
Extremely high training speed; scalable optimization &
Lower physical and visual realism \\
\midrule

\makecell[c]{LapGym \cite{bib219}} &
Laparoscopic manipulation; path planning; human-in-the-loop control &
\makecell[c]{Imitation learning;\\ reinforcement learning} &
Highly extensible; multimodal sensing support &
Limited task diversity; user-dependent fidelity \\
\midrule

\makecell[c]{SonoGym \cite{bib330}} &
Ultrasound navigation; bone reconstruction; intervention planning &
\makecell[c]{Reinforcement learning;\\ imitation learning} &
High anatomical realism for ultrasound procedures &
Modality-specific; limited generality \\
\bottomrule
\end{tabularx}
\end{table*}

\subsection{Action Datasets}\label{sec4.3}

Action datasets provide essential supervision for surgical action modeling, imitation learning, and execution-level reasoning in medical embodied AI.

Representative datasets include JIGSAWS \cite{bib317}, which combines kinematic and video data for benchmarking imitation learning of fundamental surgical skills; CholecTrack20 \cite{bib284} and m2cai16-workflow \cite{bib322}, which support multi-tool tracking and surgical phase recognition in laparoscopic cholecystectomy; and Endoscapes \cite{bib285}, which enables scene understanding and safety-aware analysis in real surgical videos. MultiBypass140 \cite{bib320} emphasizes hierarchical modeling of phases, steps, and adverse events in complex procedures, while SurgVU24 \cite{bib321} provides long-horizon robotic surgery recordings for instrument recognition and strategy modeling. Large-scale resources such as GenSurgery \cite{bib323} extend action modeling across diverse surgical types, and multimodal datasets including MM-OR \cite{bib288} and MITI \cite{bib324} support semantic action understanding and intraoperative localization through integrated sensory signals. In addition, CoPESD \cite{bib325} targets fine-grained manipulation in endoscopic submucosal dissection, enabling detailed modeling of complex surgical actions.

\subsection{Simulation Platforms and Synthetic Datasets}\label{sec4.4}
\subsubsection{Surgical Simulation Platforms}\label{sec4.4.1}

High-fidelity surgical simulation platforms provide essential infrastructure for training and evaluating medical embodied agents, particularly for reinforcement learning and imitation learning in safety-critical scenarios. Recent open-source platforms enable scalable policy learning, realistic physical interaction, and reproducible experimentation (Table~\ref{tab6}).

Representative platforms include SurRoL \cite{bib326}, which offers high-fidelity surgical interaction with collision modeling, haptics, and demonstration collection for imitation and reinforcement learning; ORBIT-Surgical \cite{bib327}, which emphasizes photorealistic rendering and GPU-parallel training for dexterous manipulation; and Surgical Gym \cite{bib328}, a fully GPU-based simulator that significantly accelerates large-scale reinforcement learning. LapGym \cite{bib219} focuses on robot-assisted laparoscopic surgery and supports multimodal perception, path planning, and human-in-the-loop learning, while SonoGym \cite{bib330} targets ultrasound-guided navigation and intervention using anatomically realistic models. Together, these platforms form a critical foundation for scalable training, safe policy optimization, and sim-to-real transfer in medical embodied AI.

\subsubsection{Synthetic Datasets}\label{sec4.4.2}

Due to limited access to real-world medical data and strict privacy constraints, synthetic datasets have become an important complement for training and evaluating medical embodied AI models. Recent synthetic resources span imaging, pathology, and structured clinical data, providing scalable and privacy-preserving alternatives.

Representative datasets include SynFundus-1M \cite{bib331}, a large-scale synthetic fundus dataset covering multiple disease categories with fine-grained anatomical quality annotations; SNOW \cite{bib332}, which provides densely annotated synthetic pathology images for nuclei segmentation in breast cancer; and COVID-19 10K/100K \cite{bib333}, synthetic electronic health record datasets generated with Synthea to model disease progression and clinical workflows. Coherent dataset \cite{bib334} further integrates synthetic multimodal data across FHIR-based clinical records, medical imaging, genomics, and physiological signals, supporting end-to-end multimodal learning and system-level evaluation. Together, these datasets enable scalable model training, benchmarking, and sim-to-real analysis under realistic privacy constraints.

\section{Challenges and Outlook}\label{sec5}

Despite recent advances, medical embodied AI still faces fundamental challenges in achieving safe, robust, and reliable deployment within complex clinical environments. This section systematically analyzes these challenges from three core perspectives—perception, decision-making, and action—and discusses promising research directions toward clinically deployable embodied AI systems.

\subsection{Challenges and Outlook in Medical Embodied Perception}\label{sec5.1}

\subsubsection{Insufficient Training Data and Perception Discrepancy}\label{sec5.1.1}
Medical embodied perception relies on large-scale, high-quality annotations for safety-critical tasks such as intraoperative navigation, instrument recognition, and tissue segmentation. Ethical and privacy constraints and the high cost of expert annotation limit data availability, leading to class imbalance and reduced domain diversity, particularly for rare pathologies and cross-institutional deployment, where domain gaps degrade robustness and increase risk.

Existing solutions include synthetic data generation, domain adaptation, semi-supervised learning, and federated learning. Synthetic data scale efficiently \cite{bib331,bib332} but fail to capture complex intraoperative variability, while domain adaptation mitigates distribution mismatch \cite{bib335} yet depends on target-domain access and struggles with unseen classes. Semi-supervised and federated learning leverage unlabeled or distributed data \cite{bib336,bib337} but remain constrained by annotation quality and domain heterogeneity. Future work should emphasize generalized perception via unified multimodal backbones, uncertainty-aware learning, and closed-loop synthetic–real co-training toward deployable clinical systems.

\subsubsection{Semantic Ambiguity and Multimodal Knowledge Fusion Difficulties}\label{sec5.1.2}
Medical embodied AI integrates heterogeneous perceptual inputs, including visual observations, clinical texts, electronic health records, voice commands, and haptic signals. Disparities in semantic granularity, spatial resolution, and temporal alignment make effective multimodal fusion challenging, often causing semantic ambiguity and intermodal conflicts in real-time clinical settings.

To bridge semantic gaps, existing approaches employ structured medical knowledge graphs and large medical language models \cite{bib154,bib179}. However, knowledge bases suffer from delayed updates and uneven coverage \cite{bib338,bib339}, while current alignment strategies rely on static mappings that struggle with dynamic contexts and evolving intraoperative semantics. Future research should therefore emphasize context-aware and adaptive semantic fusion, including graph-based multimodal reasoning, causal modeling for uncertainty-aware interpretability, and adaptive alignment via reinforcement and meta-learning, enabling a transition from task-specific sensing toward lifelong and context-aware clinical perception systems.

\subsection{Challenges and Outlook in Medical Embodied Decision-making}\label{sec5.2}
\subsubsection{Medical Reasoning Complexity and Uncertainty Modeling}\label{sec5.2.1}
Medical embodied agents operate in dynamic clinical environments with incomplete and noisy information. Intraoperative emergencies, anatomical variability, and equipment failures require multi-step reasoning under uncertainty, making transparent and interpretable decision processes essential for clinical trust.

Most existing systems rely on end-to-end policy learning or deep reinforcement learning \cite{bib24}, which perform well in controlled settings but lack explicit reasoning paths. Recent advances in stepwise medical reasoning, graph neural networks, and causal inference improve traceability and robustness; however, constructing comprehensive causal knowledge graphs remains costly and incomplete, especially for rare or individualized conditions. Future research should therefore pursue hybrid reasoning frameworks that integrate causal knowledge, expert-defined rules, and learning-based models, supported by expert-in-the-loop feedback and safety-aware assurance mechanisms. In the long term, medical embodied decision-making should evolve from static policy optimization toward explainable and human-aligned clinical reasoning agents.

\subsubsection{Lack of Mechanisms for Decision Pathway Generation and Validation}\label{sec5.2.2}
Translating reasoning outcomes into safe and executable action policies remains challenging due to the complexity and high-risk nature of clinical workflows. Although reinforcement and imitation learning show promise, most approaches lack systematic mechanisms for decision pathway validation and execution risk assessment.

Future directions include adversarial or multi-agent evaluation frameworks, digital twin–based validation platforms, and the integration of formal verification with learning-based policies to support predictable and trustworthy clinical decision-making in safety-critical clinical settings.

\subsection{Challenges and Outlook in Medical Embodied Action}\label{sec5.3}
\subsubsection{Error Sensitivity in High-Precision Action Control}\label{sec5.3.1}
High-precision medical actions impose stringent requirements on trajectory accuracy and control latency, particularly in minimally invasive procedures near sensitive anatomical structures. Errors arise from mechanical compliance, sensor latency, calibration inaccuracies, and complex tissue–instrument interactions, while most systems lack sufficient fault tolerance.

Current approaches based on visual servoing and imitation learning enable closed-loop adjustment but remain constrained by perception–control latency and limited sensing resolution. Future research should emphasize tightly coupled hardware–software frameworks, multimodal sensor fusion, and hybrid control strategies that combine model-based control with learning-based uncertainty estimation.

\subsubsection{Lack of General-Purpose Medical Simulation Platforms}\label{sec5.3.2}
Simulation platforms are critical for training and validating medical embodied actions, yet existing simulators lack sufficient physical fidelity and support for diverse procedures, rare events, and hierarchical task modeling. Limitations in tissue deformation modeling and instrument–tissue interaction hinder reliable sim-to-real transfer.

Future work should prioritize general-purpose simulation platforms with real-world closed-loop validation, integrating personalized anatomical models, procedural scene generation, and digital twin frameworks. Ultimately, medical embodied action must advance from precise but brittle execution toward adaptive, intention-aware, and safety-certified clinical actuation.

\subsection{Cross-Cutting Challenges in Safety, Ethics, and Reliable Deployment}\label{sec5.4}
\subsubsection{Safety Assurance}
Ensuring safety in medical embodied AI extends beyond improving perception accuracy or control precision \cite{bib2}. Risks may arise from the interaction of perception uncertainty, delayed decision-making, and actuation errors under dynamic clinical conditions \cite{bib16}. While simulation-based evaluation and empirical testing are widely adopted, they provide limited guarantees under rare or unforeseen scenarios \cite{bib14}. Future research should integrate runtime monitoring, fail-safe mechanisms, and human-in-the-loop supervision to support bounded-risk operation in safety-critical settings.

\subsubsection{Ethical and Regulatory Considerations}
Medical embodied AI introduces ethical challenges related to patient autonomy, informed consent, accountability, and data governance. The increasing use of multimodal clinical data and autonomous decision-making complicates responsibility attribution and regulatory compliance \cite{bib5}. Transparent decision processes, traceable action logs, and alignment with existing medical regulations are essential to ensure ethical deployment and clinical acceptance.

\subsubsection{Reliable Clinical Deployment}
Reliable deployment requires medical embodied AI systems to remain robust across institutions, patient populations, and long-term operation. Domain shifts, rare events, and system drift can degrade performance and erode clinical trust \cite{bib28}. Addressing these challenges calls for standardized benchmarks, continuous post-deployment monitoring, and shared-autonomy frameworks that allow clinicians to retain ultimate control while benefiting from intelligent assistance.

\section{Conclusion}\label{sec6}
Embodied AI introduces a transformative paradigm to healthcare, effectively bridging the critical gap between computational foundation models and the physical clinical world. This review has provided a comprehensive survey of the field, systematically analyzing the core components of perception, decision-making, and action, while cataloging representative medical applications and essential datasets. Despite the promising progress, we also highlighted the significant challenges in clinical settings. By elucidating the current landscape and identifying key bottlenecks, this work aims to serve as a foundational roadmap. It is our hope that this survey will facilitate researchers in addressing these limitations, ultimately accelerating the transition of intelligent agents from theoretical frameworks to practical, reliable assistants in real-world medical workflows.

\bibliographystyle{IEEEtran}
\bstctlcite{IEEEexample:BSTcontrol}
\bibliography{ref}

\begin{thebibliography}{100}
\providecommand{\url}[1]{#1}
\csname url@samestyle\endcsname
\providecommand{\newblock}{\relax}
\providecommand{\bibinfo}[2]{#2}
\providecommand{\BIBentrySTDinterwordspacing}{\spaceskip=0pt\relax}
\providecommand{\BIBentryALTinterwordstretchfactor}{4}
\providecommand{\BIBentryALTinterwordspacing}{\spaceskip=\fontdimen2\font plus
\BIBentryALTinterwordstretchfactor\fontdimen3\font minus \fontdimen4\font\relax}
\providecommand{\BIBforeignlanguage}[2]{{%
\expandafter\ifx\csname l@#1\endcsname\relax
\typeout{** WARNING: IEEEtran.bst: No hyphenation pattern has been}%
\typeout{** loaded for the language `#1'. Using the pattern for}%
\typeout{** the default language instead.}%
\else
\language=\csname l@#1\endcsname
\fi
#2}}
\providecommand{\BIBdecl}{\relax}
\BIBdecl

\bibitem{bib2}
C.~Varghese, E.~M. Harrison, G.~O'Grady, and E.~J. Topol, ``Artificial intelligence in surgery,'' \emph{Nat. Med.}, vol.~30, no.~5, pp. 1257--1268, 2024.

\bibitem{bib1}
F.~Isensee, P.~F. Jaeger, S.~A.~A. Kohl, J.~Petersen, and K.~H. Maier-Hein, ``{nnU-Net}: A self-configuring method for deep learning-based biomedical image segmentation,'' \emph{Nat. Methods}, vol.~18, no.~2, pp. 203--211, 2021.

\bibitem{bib3}
A.~H. Thieme, Y.~Zheng, G.~Machiraju, C.~Sadee, M.~Mittermaier, M.~Gertler, J.~L. Salinas, K.~Srinivasan, P.~Gyawali, and F.~C.-P. et~al., ``A deep-learning algorithm to classify skin lesions from mpox virus infection,'' \emph{Nat. Med.}, vol.~29, no.~3, pp. 738--747, 2023.

\bibitem{bib4}
D.~Ma, J.~Pang, M.~B. Gotway, and J.~Liang, ``A fully open {AI} foundation model applied to chest radiography,'' \emph{Nature}, pp. 1--11, 2025.

\bibitem{bib5}
F.~Liu, H.~Zhou, B.~Gu, X.~Zou, J.~Huang, J.~Wu, Y.~Li, S.~S. Chen, Y.~Hua, and P.~Z. et~al., ``Application of large language models in medicine,'' \emph{Nat. Rev. Bioeng.}, pp. 1--20, 2025.

\bibitem{bib6}
K.~Singhal, T.~Tu, J.~Gottweis, R.~Sayres, E.~Wulczyn, M.~Amin, L.~Hou, K.~Clark, S.~R. Pfohl, and H.~C.-L. et~al., ``Toward expert-level medical question answering with large language models,'' \emph{Nat. Med.}, vol.~31, no.~3, pp. 943--950, 2025.

\bibitem{bib7}
X.~Liu, H.~Liu, G.~Yang, Z.~Jiang, S.~Cui, Z.~Zhang, H.~Wang, L.~Tao, Y.~Sun, and Z.~S. et~al., ``A generalist medical language model for disease diagnosis assistance,'' \emph{Nat. Med.}, vol.~31, no.~3, pp. 932--942, 2025.

\bibitem{bib8}
Y.~Liu, W.~Chen, Y.~Bai, X.~Liang, G.~Li, W.~Gao, and L.~Lin, ``Aligning cyber space with physical world: A comprehensive survey on embodied {AI},'' \emph{IEEE/ASME Trans. Mechatronics}, 2025.

\bibitem{bib9}
J.~Li, Z.~Xu, N.~Li, K.~Zhang, G.~Xiong, M.~Sun, C.~Hou, J.~Ji, F.~Zhang, and J.~Z. et~al., ``{AI}-embodied multimodal flexible electronic robots with programmable sensing, actuating, and self-learning,'' \emph{Nat. Commun.}, vol.~16, no.~1, p. 8818, 2025.

\bibitem{bib10}
Y.~Long, A.~Lin, D.~H.~C. Kwok, L.~Zhang, Z.~Yang, K.~Shi, L.~Song, J.~Fu, H.~Lin, and W.~W. et~al., ``Surgical embodied intelligence for generalized task autonomy in laparoscopic robot-assisted surgery,'' \emph{Sci. Robot.}, vol.~10, no. 104, p. eadt3093, 2025.

\bibitem{bib11}
P.~Fiorini, K.~Y. Goldberg, Y.~Liu, and R.~H. Taylor, ``Concepts and trends in autonomy for robot-assisted surgery,'' \emph{Proc. IEEE}, vol. 110, no.~7, pp. 993--1011, 2022.

\bibitem{bib14}
T.~Yao, H.~Wang, B.~Lu, J.~Ge, Z.~Pei, M.~Kowarschik, L.~Sun, L.~Seneviratne, and P.~Qi, ``Sim-to-real learning with domain randomization for autonomous guidewire navigation in robot-assisted endovascular procedures,'' \emph{IEEE Trans. Autom. Sci. Eng.}, 2025.

\bibitem{bib15}
T.~Yao, Y.~Xu, H.~Wang, X.~Qiu, K.~Althoefer, and P.~Qi, ``Multi-agent fuzzy reinforcement learning with {LLM} for cooperative navigation of endovascular robotics,'' \emph{IEEE Trans. Fuzzy Syst.}, 2025.

\bibitem{bib16}
A.~Pore, Z.~Li, D.~Dall'Alba, A.~Hernansanz, E.~D. Momi, A.~Menciassi, A.~C. Gelpi, J.~Dankelman, P.~Fiorini, and E.~V. Poorten, ``Autonomous navigation for robot-assisted intraluminal and endovascular procedures: A systematic review,'' \emph{IEEE Trans. Robot.}, vol.~39, no.~4, pp. 2529--2548, 2023.

\bibitem{bib17}
J.~Song, K.~Yang, H.~Chen, J.~Liu, Y.~Gu, Q.~Hui, Y.~Huang, M.~Li, Z.~Zhang, and T.~C. et~al., ``{VascularPilot3D}: Toward a {3D} fully autonomous navigation for endovascular robotics,'' in \emph{Proc. IEEE Int. Conf. Robot. Autom. (ICRA)}, 2025, pp. 9318--9324.

\bibitem{bib18}
J.~Song, R.~Zhang, W.~Zhang, H.~Zhou, and M.~Ghaffari, ``{SLAM}-assisted {3D} tracking system for laparoscopic surgery,'' in \emph{Proc. IEEE Int. Conf. Robot. Autom. (ICRA)}, 2025, pp. 6868--6874.

\bibitem{bib20}
W.~Arreola, J.~J. Rivas, L.~Castrejon, and L.~E. Sucar, ``Affective embodied agent for patient assistance in virtual rehabilitation,'' \emph{IEEE Trans. Affect. Comput.}, 2025.

\bibitem{bib21}
C.~Zhang and S.~Yu, ``Virtual co-embodiment rehabilitation: An innovative method integrating virtual co-embodiment and action observation therapy in virtual reality rehabilitation,'' in \emph{Proc. Int. Conv. Rehabil. Eng. Assistive Technol. (i-CREATe)}, 2024, pp. 1--6.

\bibitem{bib22}
Z.~Jiang, X.~Huang, Z.~Wang, Y.~Liu, L.~Huang, and X.~Luo, ``Embodied conversational agents for chronic diseases: Scoping review,'' \emph{J. Med. Internet Res.}, vol.~26, p. e47134, 2024.

\bibitem{bib27}
G.~Fragapane, H.-H. Hvolby, F.~Sgarbossa, and J.~O. Strandhagen, ``Autonomous mobile robots in hospital logistics,'' in \emph{Adv. Prod. Manage. Syst.}, 2020, pp. 672--679.

\bibitem{bib28}
L.~Bernhard, P.~Schwingenschl{\"o}gl, J.~Hofmann, D.~Wilhelm, and A.~Knoll, ``Boosting the hospital by integrating mobile robotic assistance systems: A comprehensive classification of the risks to be addressed,'' \emph{Auton. Robots}, vol.~48, no.~1, p.~1, 2024.

\bibitem{bib29}
W.~Ding, Q.~Tian, Y.~Xia, Y.~Yang, Y.~Wang, and Y.~Zhang, ``Research on multirobot collaboration platform for logistic distribution of medical consumables in the operating room,'' in \emph{Proc. SPIE Conf. Biomed. Intell. Syst. (IC-BIS)}, vol. 13208, 2024, pp. 637--642.

\bibitem{bib30}
O.~Palinko, R.~Wendlandt, S.~Udby, F.~Uhing, J.~H. Fog, E.~Hansen, R.~P. Junge, D.~G. Holm, M.~Kipp, and L.~Bodenhagen, ``Interaction matters when it comes to hand disinfection using robots at hospitals,'' in \emph{Proc. Int. Conf. Social Robot.}, 2024, pp. 74--85.

\bibitem{bib31}
Y.~Liu, X.~Cao, T.~Chen, Y.~Jiang, J.~You, M.~Wu, X.~Wang, M.~Feng, Y.~Jin, and J.~Chen, ``From screens to scenes: A survey of embodied {AI} in healthcare,'' \emph{Inf. Fusion}, vol. 119, p. 103033, 2025.

\bibitem{bib34}
Z.~Zhong, ``Hierarchical frameworks for embodied medical {AI},'' \emph{ITM Web of Conferences}, vol.~80, p. 01037, 2025.

\bibitem{bib35}
S.~N. Kumar, J.~Joy, A.~J. James, and A.~Dixen, ``Health care industry use cases of embodied {AI},'' \emph{Building Embodied AI Systems}, pp. 223--239, 2025.

\bibitem{bib36}
Y.~Tian, M.~Shi, X.~Zhang, B.~Zhang, M.~Wang, and Y.~Shi, ``Assisting embodied {AI}: A survey of {3D} segmentation models for medical {CT} images,'' \emph{CCF Transactions on Pervasive Computing and Interaction}, pp. 1--22, 2025.

\bibitem{bib37}
Y.~Qiu, X.~Chen, X.~Wu, Y.~Li, P.~Xu, K.~Jin, X.~Shang, P.~Chotcomwongse, M.~He, and D.~Shi, ``Embodied artificial intelligence in ophthalmology,'' \emph{npj Digital Medicine}, vol.~8, no.~1, p. 351, 2025.

\bibitem{bib38}
J.~Liu, X.~Shi, T.~D. Nguyen, H.~Zhang, T.~Zhang, W.~Sun, Y.~Li, A.~V. Vasilakos, G.~Iacca, and A.~A.~K. et~al., ``Neural brain: A neuroscience-inspired framework for embodied agents,'' \emph{arXiv preprint arXiv:2505.07634}, 2025.

\bibitem{bib39}
H.~Liu, D.~Guo, and A.~Cangelosi, ``Embodied intelligence: A synergy of morphology, action, perception, and learning,'' \emph{ACM Comput. Surv.}, vol.~57, no.~7, pp. 1--36, 2025.

\bibitem{bib40}
G.~Paolo, J.~Gonzalez-Billandon, and B.~K{\'e}gl, ``Position: A call for embodied {AI},'' in \emph{Proc. Int. Conf. Mach. Learn. (ICML)}, 2024.

\bibitem{bib42}
B.~Wang, X.~Meng, X.~Wang, Z.~Zhu, A.~Ye, Y.~Wang, Z.~Yang, C.~Ni, G.~Huang, and X.~Wang, ``{EmbodiedDreamer}: Advancing real-to-sim-to-real transfer for policy training via embodied world modeling,'' \emph{arXiv preprint arXiv:2507.05198}, 2025.

\bibitem{bib43}
T.~Jiang, Y.~Guan, L.~Ma, J.~Xu, J.~Meng, W.~Chen, Z.~Zeng, L.~Li, D.~Wu, and R.~Chen, ``{DexSim2Real}$^{2}$: Building explicit world model for precise articulated object dexterous manipulation,'' \emph{arXiv preprint arXiv:2409.08750}, 2024.

\bibitem{bib44}
Y.~Yardi, S.~Biruduganti, and L.~Ankile, ``Bridging the sim-to-real gap: Vision encoder pre-training for visuomotor policy transfer,'' \emph{arXiv preprint arXiv:2501.16389}, 2025.

\bibitem{bib45}
G.~Liu, Y.~Deng, R.~Zhao, H.~Zhou, J.~Chen, J.~Chen, R.~Xu, Y.~Tai, and K.~Jia, ``{DexScale}: Automating data scaling for {Sim2Real} generalizable robot control,'' in \emph{Proc. Int. Conf. Mach. Learn. (ICML)}, 2025.

\bibitem{bib46}
L.~Fan, M.~Liang, Y.~Li, G.~Hua, and Y.~Wu, ``Evidential active recognition: Intelligent and prudent open-world embodied perception,'' in \emph{Proc. IEEE/CVF Conf. Comput. Vis. Pattern Recognit. (CVPR)}, 2024, pp. 16\,351--16\,361.

\bibitem{bib49}
Y.~Sun, N.~Cheng, S.~Zhang, W.~Li, L.~Yang, S.~Cui, H.~Liu, F.~Sun, J.~Zhang, and G.~D. et~al., ``Tactile data generation and applications based on visuo-tactile sensors: A review,'' \emph{Inf. Fusion}, p. 103162, 2025.

\bibitem{bib99}
W.~Jin, H.~Du, B.~Zhao, X.~Tian, B.~Shi, and G.~Yang, ``A comprehensive survey on multi-agent cooperative decision-making: Scenarios, approaches, challenges, and perspectives,'' \emph{arXiv preprint arXiv:2503.13415}, 2025.

\bibitem{bib01}
R.~Liu, W.~Wang, and Y.~Yang, ``Volumetric environment representation for vision-language navigation,'' in \emph{Proc. IEEE/CVF Conf. Comput. Vis. Pattern Recognit. (CVPR)}, 2024, pp. 16\,317--16\,328.

\bibitem{bib03}
R.~Liu, X.~Wang, W.~Wang, and Y.~Yang, ``Bird's-eye-view scene graph for vision-language navigation,'' in \emph{Proc. IEEE/CVF Int. Conf. Comput. Vis. (ICCV)}, 2023, pp. 10\,968--10\,980.

\bibitem{bib02}
R.~Liu, W.~Wang, and Y.~Yang, ``Vision-language navigation with energy-based policy,'' in \emph{Proc. Adv. Neural Inf. Process. Syst. (NeurIPS)}, 2024, pp. 108\,208--108\,230.

\bibitem{bib120}
S.~Saxena, B.~Buchanan, C.~Paxton, P.~Liu, B.~Chen, N.~Vaskevicius, L.~Palmieri, J.~Francis, and O.~Kroemer, ``{GraphEQA}: Using {3D} semantic scene graphs for real-time embodied question answering,'' \emph{arXiv preprint arXiv:2412.14480}, 2024.

\bibitem{bib149}
Y.~Lei, Y.~Fu, T.~Wang, R.~L.~J. Qiu, W.~J. Curran, T.~Liu, and X.~Yang, ``Deep learning in multi-organ segmentation,'' \emph{arXiv preprint arXiv:2001.10619}, 2020.

\bibitem{bib150}
F.~A. Ahmed, M.~Yousef, M.~A. Ahmed, H.~O. Ali, A.~Mahboob, H.~Ali, Z.~Shah, O.~Aboumarzouk, A.~Al~Ansari, and S.~Balakrishnan, ``Deep learning for surgical instrument recognition and segmentation in robotic-assisted surgeries: a systematic review,'' \emph{Artif. Intell. Rev.}, vol.~58, no.~1, p.~1, 2024.

\bibitem{bib57}
O.~Ronneberger, P.~Fischer, and T.~Brox, ``{U}-{Net}: Convolutional networks for biomedical image segmentation,'' in \emph{Proc. Int. Conf. Med. Image Comput. Comput.-Assist. Interv. (MICCAI)}, 2015, pp. 234--241.

\bibitem{bib151}
H.~Du, J.~Wang, M.~Liu, Y.~Wang, and E.~Meijering, ``Swinpa-net: Swin transformer-based multiscale feature pyramid aggregation network for medical image segmentation,'' \emph{IEEE Trans. Neural Netw. Learn. Syst.}, vol.~35, no.~4, pp. 5355--5366, 2022.

\bibitem{bib152}
M.~Islam, V.~S. Vibashan, C.~M. Lim, and H.~Ren, ``St-mtl: Spatio-temporal multitask learning model to predict scanpath while tracking instruments in robotic surgery,'' \emph{Med. Image Anal.}, vol.~67, p. 101837, 2021.

\bibitem{bib153}
E.~Colleoni, S.~Moccia, X.~Du, E.~De~Momi, and D.~Stoyanov, ``Deep learning based robotic tool detection and articulation estimation with spatio-temporal layers,'' \emph{IEEE Robot. Autom. Lett.}, vol.~4, no.~3, pp. 2714--2721, 2019.

\bibitem{bib154}
Z.~Zeng, Z.~Zhuo, X.~Jia, E.~Zhang, J.~Wu, J.~Zhang, Y.~Wang, C.~H. Low, J.~Jiang, Z.~Zheng \emph{et~al.}, ``Surgvlm: A large vision-language model and systematic evaluation benchmark for surgical intelligence,'' \emph{arXiv preprint arXiv:2506.02555}, 2025.

\bibitem{bib155}
Z.~Li, A.~Shaban, J.-G. Simard, D.~Rabindran, S.~DiMaio, and O.~Mohareri, ``A robotic {3D} perception system for operating room environment awareness,'' \emph{arXiv preprint arXiv:2003.09487}, 2020.

\bibitem{bib156}
G.~Erol, A.~G{\"u}ng{\"o}r, U.~T. Sevgi, B.~G{\"u}lsuna, Y.~Do{\u{g}}ruel, H.~Emmez, and U.~T{\"u}re, ``Creation of a microsurgical neuroanatomy laboratory and virtual operating room: a preliminary study,'' \emph{Neurosurg. Focus}, vol.~56, no.~1, p.~E6, 2024.

\bibitem{bib157}
B.~G.~A. Gerats, J.~M. Wolterink, and I.~A. M.~J. Broeders, ``{NeRF}-or: Neural radiance fields for operating room scene reconstruction from sparse-view {RGB}-{D} videos,'' \emph{Int. J. Comput. Assist. Radiol. Surg.}, vol.~20, no.~1, pp. 147--156, 2025.

\bibitem{bib158}
S.~Yang, Q.~Li, D.~Shen, B.~Gong, Q.~Dou, and Y.~Jin, ``Deform{3DGS}: Flexible deformation for fast surgical scene reconstruction with gaussian splatting,'' in \emph{Proc. Int. Conf. Med. Image Comput. Comput.-Assist. Interv. (MICCAI)}.\hskip 1em plus 0.5em minus 0.4em\relax Springer, 2024, pp. 132--142.

\bibitem{bib1580}
W.~Xie, Y.~Ye, Q.~Hong, J.~Yao, S.~Wu, R.~Zhou, X.~Dong, and X.~Guo, ``Endo-hdr: Dynamic endoscopic reconstruction with deformable 3d gaussians and hierarchical depth regularization.''\hskip 1em plus 0.5em minus 0.4em\relax Elsevier, 2025, p. 114914.

\bibitem{bib1581}
J.~Chen, X.~Zhang, M.~I. Hoque, F.~Vasconcelos, D.~Stoyanov, D.~S. Elson, and B.~Huang, ``Surgicalgs: Dynamic 3d gaussian splatting for accurate robotic-assisted surgical scene reconstruction,'' in \emph{International Conference on Medical Image Computing and Computer-Assisted Intervention}.\hskip 1em plus 0.5em minus 0.4em\relax Springer, 2025, pp. 572--582.

\bibitem{bib159}
E.~{\"O}zsoy, E.~P. {\"O}rnek, U.~Eck, T.~Czempiel, F.~Tombari, and N.~Navab, ``{4D}-or: Semantic scene graphs for {OR} domain modeling,'' in \emph{Proc. Int. Conf. Med. Image Comput. Comput.-Assist. Interv. (MICCAI)}.\hskip 1em plus 0.5em minus 0.4em\relax Springer, 2022, pp. 475--485.

\bibitem{bib160}
E.~{\"O}zsoy, T.~Czempiel, F.~Holm, C.~Pellegrini, and N.~Navab, ``Labrad-or: Lightweight memory scene graphs for accurate bimodal reasoning in dynamic operating rooms,'' in \emph{Proc. Int. Conf. Med. Image Comput. Comput.-Assist. Interv. (MICCAI)}.\hskip 1em plus 0.5em minus 0.4em\relax Springer, 2023, pp. 302--311.

\bibitem{bib161}
P.~He, Z.~Zhang, Y.~Zhang, X.~Zhao, and S.~Peng, ``{Spatial}-{ORMLLM}: Improve spatial relation understanding in the operating room with multimodal large language models,'' \emph{arXiv preprint arXiv:2508.08199}, 2025.

\bibitem{bib162}
K.~C. Demir, H.~Schieber, T.~Weise, D.~Roth, M.~May, A.~Maier, and S.~H. Yang, ``Deep learning in surgical workflow analysis: A review of phase and step recognition,'' \emph{IEEE J. Biomed. Health Inform.}, vol.~27, no.~11, pp. 5405--5417, 2023.

\bibitem{bib165}
K.~Feghoul, D.~S. Maia, M.~E. Amrani, M.~Daoudi, and A.~Amad, ``{MGRFormer}: A multimodal transformer approach for surgical gesture recognition,'' in \emph{Proc. IEEE Int. Conf. Autom. Face Gesture Recognit. (FG)}, 2024, pp. 1--10.

\bibitem{bib166}
Y.~Men, J.~Luo, Z.~Zhao, H.~Wu, F.~Luo, G.~Zhang, and M.~Yu, ``Surgical gesture recognition in open surgery based on {3D}{CNN} and {SlowFast},'' in \emph{Proc. IEEE Int. Conf. Inf. Technol. Netw. Electron. Autom. Control (ITNEC)}, 2024, pp. 429--433.

\bibitem{bib167}
L.~Ma, H.~Kang, N.~Magnenat-Thalmann, and K.~Wac, ``{TransSG}: A spatio-temporal transformer for surgical gesture recognition,'' in \emph{Proc. Comput. Graph. Int. Conf.}, 2024, pp. 151--165.

\bibitem{bib168}
B.~Jia, W.~Wang, X.~Tian, and X.~Wang, ``{STANet}: A surgical gesture recognition method based on spatiotemporal fusion,'' \emph{Ann. N. Y. Acad. Sci.}, 2025.

\bibitem{bib169}
S.~Cristina, V.~Despotovic, R.~P{\'e}rez-Rodr{\'\i}guez, and S.~Aleksic, ``Audio- and video-based human activity recognition systems in healthcare,'' \emph{IEEE Access}, vol.~12, pp. 8230--8245, 2024.

\bibitem{bib170}
B.~V. Amsterdam, I.~Funke, E.~Edwards, S.~Speidel, J.~Collins, A.~Sridhar, J.~Kelly, M.~J. Clarkson, and D.~Stoyanov, ``Gesture recognition in robotic surgery with multimodal attention,'' \emph{IEEE Trans. Med. Imaging}, vol.~41, no.~7, pp. 1677--1687, 2022.

\bibitem{bib171}
S.~Li and W.~Deng, ``Deep facial expression recognition: A survey,'' \emph{IEEE Trans. Affect. Comput.}, vol.~13, no.~3, pp. 1195--1215, 2020.

\bibitem{bib172}
Y.~Li, J.~Wei, Y.~Liu, J.~Kauttonen, and G.~Zhao, ``Deep learning for micro-expression recognition: A survey,'' \emph{IEEE Trans. Affect. Comput.}, vol.~13, no.~4, pp. 2028--2046, 2022.

\bibitem{bib173}
L.~Zhang, Y.~Qian, O.~Arandjelovi{\'c}, T.~Zhu, and H.~Xiao, ``Multimodal latent emotion recognition from micro-expression and physiological signals,'' \emph{Pattern Recognit.}, p. 111963, 2025.

\bibitem{bib174}
F.~Zhang, Y.~Liu, X.~Yu, Z.~Wang, Q.~Zhang, J.~Wang, and Q.~Zhang, ``Towards facial micro-expression detection and classification using modified multimodal ensemble learning,'' \emph{Inf. Fusion}, vol. 115, p. 102735, 2025.

\bibitem{bib175}
J.~Ye, Y.~Yu, L.~Lu, H.~Wang, Y.~Zheng, Y.~Liu, and Q.~Wang, ``{DEP}-former: Multimodal depression recognition based on facial expressions and audio features via emotional changes,'' \emph{IEEE Trans. Circuits Syst. Video Technol.}, 2024.

\bibitem{bib176}
M.~Khan, W.~Gueaieb, A.~E. Saddik, and S.~Kwon, ``{MSER}: Multimodal speech emotion recognition using cross-attention with deep fusion,'' \emph{Expert Syst. Appl.}, vol. 245, p. 122946, 2024.

\bibitem{bib177}
H.~Gao, Z.~Cai, X.~Wang, M.~Wu, and C.~Liu, ``Multimodal fusion of behavioral and physiological signals for enhanced emotion recognition via feature decoupling and knowledge transfer,'' \emph{IEEE J. Biomed. Health Inform.}, 2025.

\bibitem{bib178}
P.~S. Kumar, P.~K. Govarthan, A.~A.~S. Gadda, N.~Ganapathy, and J.~F.~A. Ronickom, ``Deep learning-based automated emotion recognition using multimodal physiological signals and time-frequency methods,'' \emph{IEEE Trans. Instrum. Meas.}, vol.~73, pp. 1--12, 2024.

\bibitem{bib179}
J.~Pan, C.~Liu, J.~Wu, F.~Liu, J.~Zhu, H.~B. Li, C.~Chen, C.~Ouyang, and D.~Rueckert, ``{MedVLM}-{R1}: Incentivizing medical reasoning capability of vision-language models via reinforcement learning,'' in \emph{Proc. Int. Conf. Med. Image Comput. Comput.-Assist. Interv. (MICCAI)}, 2025, pp. 337--347.

\bibitem{bib180}
Y.~Zhang, M.~Wang, Y.~Wu, P.~Tiwari, Q.~Li, B.~Wang, and J.~Qin, ``{DialogueLLM}: Context- and emotion-knowledge-tuned large language models for emotion recognition in conversations,'' \emph{arXiv preprint arXiv:2310.11374}, 2023.

\bibitem{bib181}
C.~Garcia-Vidal, G.~Sanjuan, P.~Puerta-Alcalde, E.~Moreno-Garc{\'\i}a, and A.~Soriano, ``Artificial intelligence to support clinical decision-making processes,'' \emph{EBioMedicine}, vol.~46, pp. 27--29, 2019.

\bibitem{bib182}
T.~J. Loftus, P.~J. Tighe, A.~C. Filiberto, P.~A. Efron, S.~C. Brakenridge, A.~M. Mohr, P.~Rashidi, G.~R. Upchurch, and A.~Bihorac, ``Artificial intelligence and surgical decision-making,'' \emph{JAMA Surg.}, vol. 155, no.~2, pp. 148--158, 2020.

\bibitem{bib183}
F.~G. Mangano, O.~Admakin, H.~Lerner, and C.~Mangano, ``Artificial intelligence and augmented reality for guided implant surgery planning: A proof of concept,'' \emph{J. Dent.}, vol. 133, p. 104485, 2023.

\bibitem{bib184}
E.~Aspland, D.~Gartner, and P.~Harper, ``Clinical pathway modelling: A literature review,'' \emph{Health Syst.}, vol.~10, no.~1, pp. 1--23, 2021.

\bibitem{bib185}
X.~Gao, Y.~Jin, Y.~Long, Q.~Dou, and P.-A. Heng, ``{Trans}-{SVNet}: Accurate phase recognition from surgical videos via hybrid embedding aggregation transformer,'' in \emph{Proc. Int. Conf. Med. Image Comput. Comput.-Assist. Interv. (MICCAI)}, 2021, pp. 593--603.

\bibitem{bib186}
T.~Czempiel, M.~Paschali, M.~Keicher, W.~Simson, H.~Feussner, S.~T. Kim, and N.~Navab, ``{TeCNO}: Surgical phase recognition with multi-stage temporal convolutional networks,'' in \emph{Proc. Int. Conf. Med. Image Comput. Comput.-Assist. Interv. (MICCAI)}, 2020, pp. 343--352.

\bibitem{bib187}
A.~Kadkhodamohammadi, I.~Luengo, and D.~Stoyanov, ``{PATG}: Position-aware temporal graph networks for surgical phase recognition on laparoscopic videos,'' \emph{Int. J. Comput. Assist. Radiol. Surg.}, vol.~17, no.~5, pp. 849--856, 2022.

\bibitem{bib188}
F.~X. Zhang, N.~A. Moubayed, and H.~P.~H. Shum, ``Towards graph representation learning-based surgical workflow anticipation,'' in \emph{Proc. IEEE-EMBS Int. Conf. Biomed. Health Inform. (BHI)}, 2022, pp. 1--4.

\bibitem{bib189}
C.~Li, C.~Wong, S.~Zhang, N.~Usuyama, H.~Liu, J.~Yang, T.~Naumann, H.~Poon, and J.~Gao, ``{LLaVA}-{Med}: Training a large language-and-vision assistant for biomedicine in one day,'' \emph{Adv. Neural Inf. Process. Syst.}, vol.~36, pp. 28\,541--28\,564, 2023.

\bibitem{bib194}
H.~K. Gumprecht, D.~C. Widenka, and C.~B. Lumenta, ``{BrainLab} {VectorVision} neuronavigation system: Technology and clinical experiences in 131 cases,'' \emph{Neurosurgery}, vol.~44, no.~1, pp. 97--104, 1999.

\bibitem{bib195}
P.~Liu, C.~Li, C.~Xiao, Z.~Zhang, J.~Ma, J.~Gao, P.~Shao, I.~Valerio, T.~M. Pawlik, and C.~D. et~al., ``A wearable augmented reality navigation system for surgical telementoring based on {Microsoft} {HoloLens},'' \emph{Ann. Biomed. Eng.}, vol.~49, no.~1, pp. 287--298, 2021.

\bibitem{bib196}
A.~Li, J.~Han, Y.~Zhao, M.~Q.-H. Meng, and L.~Liu, ``{RL}-{USRegi}: Autonomous ultrasound registration for radiation-free spinal surgical navigation using reinforcement learning,'' \emph{IEEE Trans. Autom. Sci. Eng.}, 2025.

\bibitem{bib197}
H.~Robertshaw, L.~Karstensen, B.~Jackson, A.~Granados, and T.~C. Booth, ``Autonomous navigation of catheters and guidewires in mechanical thrombectomy using inverse reinforcement learning,'' \emph{Int. J. Comput. Assist. Radiol. Surg.}, vol.~19, no.~8, pp. 1569--1578, 2024.

\bibitem{bib198}
G.~Zhou, Y.~Hong, and Q.~Wu, ``{NavGPT}: Explicit reasoning in vision-and-language navigation with large language models,'' in \emph{Proc. AAAI Conf. Artif. Intell.}, vol.~38, no.~7, 2024, pp. 7641--7649.

\bibitem{bib199}
G.~Zhou, Y.~Hong, Z.~Wang, X.~E. Wang, and Q.~Wu, ``{NavGPT}-2: Unleashing navigational reasoning capability for large vision-language models,'' in \emph{Proc. Eur. Conf. Comput. Vis. (ECCV)}, 2024, pp. 260--278.

\bibitem{bib04}
S.~Fan, R.~Liu, W.~Wang, and Y.~Yang, ``Navigation instruction generation with {BEV} perception and large language models,'' in \emph{Proc. Eur. Conf. Comput. Vis. (ECCV)}, 2024, pp. 368--387.

\bibitem{bib05}
{S. Fan, R. Liu, W. Wang, and Y. Yang}, ``Scene map-based prompt tuning for navigation instruction generation,'' in \emph{Proc. IEEE/CVF Conf. Comput. Vis. Pattern Recognit. (CVPR)}, 2025, pp. 6898--6908.

\bibitem{bib201}
Z.~Lin, D.~Zhang, Q.~Tao, D.~Shi, G.~Haffari, Q.~Wu, M.~He, and Z.~Ge, ``Medical visual question answering: A survey,'' \emph{Artif. Intell. Med.}, vol. 143, p. 102611, 2023.

\bibitem{bib202}
Q.~Jin, Z.~Yuan, G.~Xiong, Q.~Yu, H.~Ying, C.~Tan, M.~Chen, S.~Huang, X.~Liu, and S.~Yu, ``Biomedical question answering: A survey of approaches and challenges,'' \emph{ACM Comput. Surv.}, vol.~55, no.~2, pp. 1--36, 2022.

\bibitem{bib203}
A.~W. Rosen, I.~Ose, M.~G{\"o}genur, L.~P.~K. Andersen, R.~D. Bojesen, R.~P. Vogelsang, M.~H. Rose, P.~W. Steenfos, L.~B. Hansen, and H.~S.~S. et~al., ``Clinical implementation of an {AI}-based prediction model for decision support for patients undergoing colorectal cancer surgery,'' \emph{Nat. Med.}, pp. 1--12, 2025.

\bibitem{bib204}
A.~{\c{S}}. {\c{C}}ift{\c{c}}i and A.~H. Acar, ``Artificial intelligence-based chatbot assistance in clinical decision-making for medically complex patients in oral surgery: A comparative study,'' \emph{BMC Oral Health}, vol.~25, no.~1, p. 351, 2025.

\bibitem{bib205}
A.~Patil, V.~Patil, S.~Sankpal, T.~S. Patankar, and H.~Bhute, ``Multimodal decision support system for improved diagnosis and healthcare decision making,'' \emph{J. Biol. Health Sci.}, 2025.

\bibitem{bib206}
L.~Gong, J.~Yang, S.~Han, and Y.~Ji, ``{MedBLIP}: A multimodal method of medical question answering based on fine-tuning large language models,'' \emph{Comput. Med. Imaging Graph.}, p. 102581, 2025.

\bibitem{bib207}
S.~Schmidgall, J.~D. Opfermann, J.~W. Kim, and A.~Krieger, ``Will your next surgeon be a robot? autonomy and {AI} in robotic surgery,'' \emph{Sci. Robot.}, vol.~10, no. 104, p. eadt0187, 2025.

\bibitem{bib208}
A.~Attanasio, B.~Scaglioni, E.~D. Momi, P.~Fiorini, and P.~Valdastri, ``Autonomy in surgical robotics,'' \emph{Annu. Rev. Control Robot. Auton. Syst.}, vol.~4, no.~1, pp. 651--679, 2021.

\bibitem{bib209}
A.~Peloso, R.~Damiano, X.~Zhang, A.~Bicchi, E.~Votta, and E.~D. Momi, ``Imitation learning for path planning in cardiac percutaneous interventions,'' \emph{IEEE Trans. Biomed. Eng.}, 2025.

\bibitem{bib210}
J.~W. Kim, T.~Z. Zhao, S.~Schmidgall, A.~Deguet, M.~Kobilarov, C.~Finn, and A.~Krieger, ``Surgical robot transformer ({SRT}): Imitation learning for surgical tasks,'' \emph{arXiv preprint arXiv:2407.12998}, 2024.

\bibitem{bib211}
S.~Paradis, M.~Hwang, B.~Thananjeyan, J.~Ichnowski, D.~Seita, D.~Fer, T.~Low, J.~E. Gonzalez, and K.~Goldberg, ``Intermittent visual servoing: Efficiently learning policies robust to instrument changes for high-precision surgical manipulation,'' in \emph{Proc. IEEE Int. Conf. Robot. Autom. (ICRA)}, 2021, pp. 7166--7173.

\bibitem{bib212}
M.~Moghani, N.~Nelson, M.~Ghanem, A.~Diaz-Pinto, K.~Hari, M.~Azizian, K.~Goldberg, S.~Huver, and A.~Garg, ``{SuFIA}-{BC}: Generating high-quality demonstration data for visuomotor policy learning in surgical subtasks,'' \emph{arXiv preprint arXiv:2504.14857}, 2025.

\bibitem{bib213}
A.~Segato, M.~D. Marzo, S.~Zucchelli, S.~Galvan, R.~Secoli, and E.~D. Momi, ``Inverse reinforcement learning intra-operative path planning for steerable needles,'' \emph{IEEE Trans. Biomed. Eng.}, vol.~69, no.~6, pp. 1995--2005, 2021.

\bibitem{bib214}
W.~Chi, G.~Dagnino, T.~M.~Y. Kwok, A.~Nguyen, D.~Kundrat, M.~E. M.~K. Abdelaziz, C.~Riga, C.~Bicknell, and G.-Z. Yang, ``Collaborative robot-assisted endovascular catheterization with generative adversarial imitation learning,'' in \emph{Proc. IEEE Int. Conf. Robot. Autom. (ICRA)}, 2020, pp. 2414--2420.

\bibitem{bib215}
A.~Gomaa, B.~Mahdy, N.~Kleer, and A.~Kr{\"u}ger, ``Towards a surgeon-in-the-loop ophthalmic robotic apprentice using reinforcement and imitation learning,'' in \emph{Proc. IEEE/RSJ Int. Conf. Intell. Robots Syst. (IROS)}, 2024, pp. 6939--6946.

\bibitem{bib216}
B.~Li, R.~Wei, J.~Xu, B.~Lu, C.~H. Yee, C.~F. Ng, P.-A. Heng, Q.~Dou, and Y.-H. Liu, ``{3D} perception-based imitation learning under limited demonstration for laparoscope control in robotic surgery,'' in \emph{Proc. IEEE Int. Conf. Robot. Autom. (ICRA)}, 2022, pp. 7664--7670.

\bibitem{bib217}
C.~Yu, J.~Liu, S.~Nemati, and G.~Yin, ``Reinforcement learning in healthcare: A survey,'' \emph{ACM Comput. Surv.}, vol.~55, no.~1, pp. 1--36, 2021.

\bibitem{bib218}
V.~M. Varier, D.~K. Rajamani, N.~Goldfarb, F.~Tavakkolmoghaddam, A.~Munawar, and G.~S. Fischer, ``Collaborative suturing: A reinforcement learning approach to automate hand-off tasks in suturing for surgical robots,'' in \emph{Proc. IEEE Int. Conf. Robot Human Interact. Commun. (RO-MAN)}, 2020, pp. 1380--1386.

\bibitem{bib219}
P.~M. Scheikl, B.~Gyenes, R.~Younis, C.~Haas, G.~Neumann, M.~Wagner, and F.~Mathis-Ullrich, ``{LapGym}: An open-source framework for reinforcement learning in robot-assisted laparoscopic surgery,'' \emph{J. Mach. Learn. Res.}, vol.~24, no. 368, pp. 1--42, 2023.

\bibitem{bib220}
F.~Meng, S.~Guo, W.~Zhou, and Z.~Chen, ``Evaluation of an autonomous navigation method for vascular interventional surgery in virtual environments,'' in \emph{Proc. IEEE Int. Conf. Mechatronics Autom. (ICMA)}, 2022, pp. 1599--1604.

\bibitem{bib221}
J.~Liu, A.~Andres, Y.~Jiang, X.~Luo, W.~Shu, and S.~A. Tsaftaris, ``Surgical task automation using actor--critic frameworks and self-supervised imitation learning,'' \emph{arXiv preprint}, 2024, arXiv:2409.02724.

\bibitem{bib225}
H.~Li, X.-H. Zhou, X.-L. Xie, S.-Q. Liu, Z.-Q. Feng, and Z.-G. Hou, ``{CASOG}: Conservative actor--critic with smooth gradient for skill learning in robot-assisted intervention,'' \emph{IEEE Trans. Ind. Electron.}, vol.~71, no.~7, pp. 7722--7731, 2023.

\bibitem{bib222}
Z.~Min, J.~Lai, and H.~Ren, ``Innovating robot-assisted surgery through large vision models,'' \emph{Nat. Rev. Electr. Eng.}, pp. 1--14, 2025.

\bibitem{bib223}
S.~Schmidgall, J.~Cho, C.~Zakka, and W.~Hiesinger, ``{GP-VLS}: A general-purpose vision--language model for surgery,'' \emph{arXiv preprint}, 2024, arXiv:2407.19305.

\bibitem{bib224}
A.~Moglia, K.~Georgiou, P.~Cerveri, L.~Mainardi, R.~M. Satava, and A.~Cuschieri, ``Large language models in healthcare: From a systematic review on medical examinations to a comparative analysis on fundamentals of robotic surgery online test,'' \emph{Artif. Intell. Rev.}, vol.~57, no.~9, p. 231, 2024.

\bibitem{bib226}
L.~Seenivasan, M.~Islam, G.~Kannan, and H.~Ren, ``{SurgicalGPT}: End-to-end language--vision {GPT} for visual question answering in surgery,'' in \emph{Proc. Int. Conf. Med. Image Comput. Comput.-Assist. Interv. (MICCAI)}, 2023, pp. 281--290.

\bibitem{bib227}
M.~Moor, Q.~Huang, S.~Wu, M.~Yasunaga, Y.~Dalmia, J.~Leskovec, C.~Zakka, E.~P. Reis, and P.~Rajpurkar, ``Med-flamingo: A multimodal medical few-shot learner,'' in \emph{Proc. Mach. Learn. Health (ML4H)}, 2023, pp. 353--367.

\bibitem{bib228}
S.~Li, J.~Wang, R.~Dai, W.~Ma, W.~Y. Ng, Y.~Hu, and Z.~Li, ``{RoboNurse-VLA}: Robotic scrub nurse system based on vision--language--action model,'' \emph{arXiv preprint}, 2024, arXiv:2409.19590.

\bibitem{bib229}
C.~D'Ettorre, A.~Mariani, A.~Stilli, F.~Rodriguez~y Baena, P.~Valdastri, A.~Deguet, P.~Kazanzides, R.~H. Taylor, G.~S. Fischer, S.~P. DiMaio \emph{et~al.}, ``Accelerating surgical robotics research: A review of 10 years with the da vinci research kit,'' \emph{IEEE Robot. Autom. Mag.}, vol.~28, no.~4, pp. 56--78, 2021.

\bibitem{bib232}
K.~Tadano and K.~Kawashima, ``A pneumatic laparoscope holder controlled by head movement,'' \emph{Int. J. Med. Robot. Comput. Assist. Surg.}, vol.~11, no.~3, pp. 331--340, 2015.

\bibitem{bib233}
L.~Chenin, J.~Peltier, and M.~Lefranc, ``Minimally invasive transforaminal lumbar interbody fusion with the {ROSA} spine robot and intraoperative flat-panel {CT} guidance,'' \emph{Acta Neurochir.}, vol. 158, no.~6, pp. 1125--1128, 2016.

\bibitem{bib234}
T.~L. Edwards, K.~Xue, H.~C.~M. Meenink, M.~J. Beelen, G.~J.~L. Naus, M.~P. Simunovic, M.~Latasiewicz, A.~D. Farmery, M.~D. De~Smet, and R.~E. MacLaren, ``First-in-human study of the safety and viability of intraocular robotic surgery,'' \emph{Nat. Biomed. Eng.}, vol.~2, no.~9, pp. 649--656, 2018.

\bibitem{bib235}
W.~L. Bargar, A.~Bauer, and M.~B{\"o}rner, ``Primary and revision total hip replacement using the {ROBODOC}{\textregistered} system,'' \emph{Clin. Orthop. Relat. Res.}, vol. 354, pp. 82--91, 1998.

\bibitem{bib236}
N.~Dastagir, D.~Obed, M.~Tamulevicius, K.~Dastagir, and P.~M. Vogt, ``The use of the {Symani} surgical system{\textregistered} in emergency hand trauma care,'' \emph{Surg. Innov.}, vol.~31, no.~5, pp. 460--465, 2024.

\bibitem{bib237}
W.~Kilby, J.~R. Dooley, G.~Kuduvalli, S.~Sayeh, and C.~R. Maurer, ``The {CyberKnife}{\textregistered} robotic radiosurgery system in 2010,'' \emph{Technol. Cancer Res. Treat.}, vol.~9, no.~5, pp. 433--452, 2010.

\bibitem{bib238}
C.~F. Graetzel, A.~Sheehy, and D.~P. Noonan, ``Robotic bronchoscopy drive mode of the auris monarch platform,'' in \emph{Proc. IEEE Int. Conf. Robot. Autom. (ICRA)}, 2019, pp. 3895--3901.

\bibitem{bib239}
M.~Remacle, V.~M.~N. Prasad, G.~Lawson, L.~Plisson, V.~Bachy, and S.~Van~der Vorst, ``Transoral robotic surgery ({TORS}) with the medrobotics flex{\texttrademark} system: First surgical application on humans,'' \emph{Eur. Arch. Oto-Rhino-Laryngol.}, vol. 272, no.~6, pp. 1451--1455, 2015.

\bibitem{bib240}
G.~W. Britz, S.~S. Panesar, P.~Falb, J.~Tomas, V.~Desai, and A.~Lumsden, ``Neuroendovascular-specific engineering modifications to the {CorPath GRX} robotic system,'' \emph{J. Neurosurg.}, vol. 133, no.~6, pp. 1830--1836, 2019.

\bibitem{bib241}
T.~Shibata, ``Therapeutic seal robot as biofeedback medical device: Qualitative and quantitative evaluations of robot therapy in dementia care,'' \emph{Proc. IEEE}, vol. 100, no.~8, pp. 2527--2538, 2012.

\bibitem{bib242}
K.~Tanaka, H.~Makino, K.~Nakamura, A.~Nakamura, M.~Hayakawa, H.~Uchida, M.~Kasahara, H.~Kato, and T.~Igarashi, ``Pilot study of group robot intervention on pediatric inpatients and their caregivers using {New Aibo},'' \emph{Eur. J. Pediatr.}, vol. 181, no.~3, pp. 1055--1061, 2022.

\bibitem{bib243}
A.~K. Pandey and R.~Gelin, ``A mass-produced sociable humanoid robot: {Pepper}, the first machine of its kind,'' \emph{IEEE Robot. Autom. Mag.}, vol.~25, no.~3, pp. 40--48, 2018.

\bibitem{bib244}
E.~Broadbent, K.~Loveys, G.~Ilan, G.~Chen, M.~M. Chilukuri, S.~G. Boardman, P.~M. Doraiswamy, and D.~Skuler, ``{ElliQ}: An {AI}-driven social robot to alleviate loneliness: Progress and lessons learned,'' \emph{J. Aging Res. Lifestyle}, vol.~13, pp. 22--28, 2024.

\bibitem{bib245}
A.~Meghdari, A.~Shariati, M.~Alemi, G.~R. Vossoughi, A.~Eydi, E.~Ahmadi, B.~Mozafari, A.~Amoozandeh~Nobaveh, and R.~Tahami, ``{ARASH}: A social robot buddy to support children with cancer in a hospital environment,'' \emph{Proc. Inst. Mech. Eng. H}, vol. 232, no.~6, pp. 605--618, 2018.

\bibitem{bib246}
Z.~H. Khan, A.~Siddique, and C.~W. Lee, ``Robotics utilization for healthcare digitization in global {COVID-19} management,'' \emph{Int. J. Environ. Res. Public Health}, vol.~17, no.~11, p. 3819, 2020.

\bibitem{bib247}
J.~Gonz{\'a}lez-Jim{\'e}nez, C.~Galindo, and J.~R. Ruiz-Sarmiento, ``Technical improvements of the {Giraff} telepresence robot based on users' evaluation,'' in \emph{Proc. IEEE Int. Symp. Robot Human Interact. Commun. (RO-MAN)}, 2012, pp. 827--832.

\bibitem{bib248}
K.~Ogawa, S.~Nishio, K.~Koda, K.~Taura, T.~Minato, C.~T. Ishii, and H.~Ishiguro, ``{Telenoid}: Tele-presence android for communication,'' in \emph{ACM SIGGRAPH Emerging Technologies}, 2011, p.~1.

\bibitem{bib249}
R.~E. Clark, D.~F. Feldon, J.~J.~G. van Merri{\"e}nboer, K.~A. Yates, and S.~Early, ``Cognitive task analysis,'' in \emph{Handbook of Research on Educational Communications and Technology}, 2008, pp. 577--593.

\bibitem{bib250}
M.~Y. Kolesnyk, ``First experience of using the {Body Interact} simulation platform in intern attestation,'' 2020.

\bibitem{bib251}
K.~Gallagher, S.~Bahadori, J.~Antonis, T.~Immins, T.~W. Wainwright, and R.~Middleton, ``Validation of the hip arthroscopy module of the {VirtaMed} virtual reality arthroscopy trainer,'' \emph{Surg. Technol. Int.}, vol.~34, pp. 430--436, 2019.

\bibitem{bib252}
V.~A. Vasilev and S.~N. Kondrichina, ``Possibilities for using the {Vimedix 3.2} virtual simulator to train ultrasound specialists,'' \emph{Digit. Diagn.}, vol.~5, no.~1, pp. 41--52, 2024.

\bibitem{bib253}
M.~Keller, S.~Zuffi, M.~J. Black, and S.~Pujades, ``{OSSO}: Obtaining skeletal shape from outside,'' in \emph{Proc. IEEE/CVF Conf. Comput. Vis. Pattern Recognit. (CVPR)}, 2022, pp. 20\,492--20\,501.

\bibitem{bib254}
J.~Lilly, ``{3D Organon VR Anatomy},'' \emph{J. Med. Libr. Assoc.}, vol. 110, no.~2, p. 276, 2022.

\bibitem{bib255}
``{Teladoc Health},'' Online, 2025, accessed: Nov. 14, 2025. Available: \url{https://www.teladochealth.com}.

\bibitem{bib256}
L.~Klingensmith and L.~Knodel, ``Mercy virtual nursing: An innovative care delivery model,'' \emph{Nurse Leader}, vol.~14, no.~4, pp. 275--279, 2016.

\bibitem{bib257}
D.~Guo, W.~Liu, X.~Zhang, M.~Zhao, B.~Zhu, T.~Hou, and H.~He, ``Duck egg white--derived peptide {VSEE} regulates bone and lipid metabolism via {Wnt}/$\beta$-catenin signaling and gut microbiota,'' \emph{Mol. Nutr. Food Res.}, vol.~63, no.~24, p. 1900525, 2019.

\bibitem{bib258}
E.~D. Kirby, B.~Beyst, J.~Beyst, S.~M. Brodie, and R.~C.~N. D'Arcy, ``A retrospective observational study of real-world clinical data from the cognitive function development therapy program,'' \emph{Front. Hum. Neurosci.}, vol.~18, p. 1508815, 2024.

\bibitem{bib230}
C.~Freschi, V.~Ferrari, F.~Melfi, M.~Ferrari, F.~Mosca, and A.~Cuschieri, ``Technical review of the da vinci surgical telemanipulator,'' \emph{Int. J. Med. Robot. Comput. Assist. Surg.}, vol.~9, no.~4, pp. 396--406, 2013.

\bibitem{bib231}
{Intuitive Surgical}, ``da vinci surgical system,'' Online, 2013, available: \url{http://www.intusurg.com/html/davinci.html}.

\bibitem{bib268}
A.~Sekuboyina, M.~E. Husseini, A.~Bayat, M.~L{\"o}ffler, H.~Liebl, H.~Li, G.~Tetteh, J.~Kuka{\v{c}}ka, C.~Payer, D.~{\v{S}}tern \emph{et~al.}, ``{VerSe}: A vertebrae labelling and segmentation benchmark for multi-detector {CT},'' \emph{Med. Image Anal.}, vol.~73, p. 102166, 2021.

\bibitem{bib269}
J.~W. van~der Graaf, M.~L. van Hooff, C.~F.~M. Buckens, M.~Rutten, J.~L.~C. van Susante, R.~J. Kroeze, M.~de~Kleuver, B.~van Ginneken, and N.~Lessmann, ``Lumbar spine segmentation in {MR} images: A dataset and public benchmark,'' \emph{Sci. Data}, vol.~11, no.~1, p. 264, 2024.

\bibitem{bib270}
Y.~Deng, C.~Wang, Y.~Hui, Q.~Li, J.~Li, S.~Luo, M.~Sun, Q.~Quan, S.~Yang, Y.~Hao \emph{et~al.}, ``{CTSpine1K}: A large-scale dataset for spinal vertebrae segmentation in computed tomography,'' \emph{arXiv preprint}, 2021, arXiv:2105.14711.

\bibitem{bib271}
P.~Liu, H.~Han, Y.~Du, H.~Zhu, Y.~Li, F.~Gu, H.~Xiao, J.~Li, C.~Zhao, L.~Xiao \emph{et~al.}, ``Deep learning for pelvic bone segmentation: Large-scale {CT} datasets and baseline models,'' \emph{Int. J. Comput. Assist. Radiol. Surg.}, vol.~16, no.~5, pp. 749--756, 2021.

\bibitem{bib272}
D.~Gutman, N.~C.~F. Codella, E.~Celebi, B.~Helba, M.~Marchetti, N.~Mishra, and A.~Halpern, ``{ISIC} 2016 challenge: Skin lesion analysis toward melanoma detection,'' \emph{arXiv preprint}, 2016, arXiv:1605.01397.

\bibitem{bib274}
T.~Mendon{\c{c}}a, M.~E. Celebi, T.~Mendonca, and J.~Marques, ``{PH$^2$}: A public database for dermoscopic image analysis,'' \emph{Dermoscopic Image Anal.}, vol.~2, 2015.

\bibitem{bib277}
J.~Wasserthal, H.-C. Breit, M.~T. Meyer, M.~Pradella, D.~Hinck, A.~W. Sauter, T.~Heye, D.~T. Boll, J.~Cyriac, S.~Yang \emph{et~al.}, ``{TotalSegmentator}: Robust segmentation of 104 anatomical structures in {CT} images,'' \emph{Radiol. Artif. Intell.}, vol.~5, no.~5, p. e230024, 2023.

\bibitem{bib278}
M.~J.~J. de~Grauw, E.~T. Scholten, E.~J. Smit, M.~J. C.~M. Rutten, M.~Prokop, B.~van Ginneken, and A.~Hering, ``{ULS23} challenge: A benchmark for universal {3D} lesion segmentation in {CT},'' \emph{Med. Image Anal.}, p. 103525, 2025.

\bibitem{bib279}
S.~Gatidis, T.~Hepp, M.~Fr{\"u}h, C.~La~Foug{\`e}re, K.~Nikolaou, C.~Pfannenberg, B.~Sch{\"o}lkopf, T.~K{\"u}stner, C.~Cyran, and D.~Rubin, ``A whole-body {FDG-PET/CT} dataset with manually annotated tumor lesions,'' \emph{Sci. Data}, vol.~9, no.~1, p. 601, 2022.

\bibitem{bib280}
M.~Allan, A.~Shvets, T.~Kurmann, Z.~Zhang, R.~Duggal, Y.-H. Su, N.~Rieke, I.~Laina, N.~Kalavakonda, S.~Bodenstedt \emph{et~al.}, ``The 2017 robotic instrument segmentation challenge,'' \emph{arXiv preprint}, 2019, arXiv:1902.06426.

\bibitem{bib281}
S.~Lin, F.~Qin, Y.~Li, R.~A. Bly, K.~S. Moe, and B.~Hannaford, ``{LC-GAN}: Image-to-image translation based on generative adversarial network for endoscopic images,'' in \emph{Proc. IEEE/RSJ Int. Conf. Intell. Robots Syst. (IROS)}, 2020, pp. 2914--2920.

\bibitem{bib282}
H.~Ding, Y.~Zhang, T.~Lu, R.~Liang, H.~Shu, L.~Seenivasan, Y.~Long, Q.~Dou, C.~Gao, Y.~Leng \emph{et~al.}, ``Segstrong-c: Segmenting surgical tools robustly on non-adversarial generated corruptions---an {EndoVis}'24 challenge,'' \emph{arXiv preprint arXiv:2407.11906}, 2024.

\bibitem{bib283}
A.~P. Twinanda, S.~Shehata, D.~Mutter, J.~Marescaux, M.~De~Mathelin, and N.~Padoy, ``Endonet: A deep architecture for recognition tasks on laparoscopic videos,'' \emph{IEEE Trans. Med. Imaging}, vol.~36, no.~1, pp. 86--97, 2016.

\bibitem{bib284}
C.~I. Nwoye, K.~Elgohary, A.~Srinivas, F.~Zaid, J.~L. Lavanchy, and N.~Padoy, ``Cholectrack20: A multi-perspective tracking dataset for surgical tools,'' in \emph{Proc. IEEE/CVF Conf. Comput. Vis. Pattern Recognit. (CVPR)}, 2025, pp. 8942--8952.

\bibitem{bib285}
A.~Murali, D.~Alapatt, P.~Mascagni, A.~Vardazaryan, A.~Garcia, N.~Okamoto, G.~Costamagna, D.~Mutter, J.~Marescaux, B.~Dallemagne \emph{et~al.}, ``The endoscapes dataset for surgical scene segmentation, object detection, and critical view of safety assessment: Official splits and benchmark,'' \emph{arXiv preprint arXiv:2312.12429}, 2023.

\bibitem{bib286}
S.~Maqbool, A.~Riaz, H.~Sajid, and O.~Hasan, ``m2caiseg: Semantic segmentation of laparoscopic images using convolutional neural networks,'' \emph{arXiv preprint arXiv:2008.10134}, 2020.

\bibitem{bib288}
E.~{\"O}zsoy, C.~Pellegrini, T.~Czempiel, F.~Tristram, K.~Yuan, D.~Bani-Harouni, U.~Eck, B.~Busam, M.~Keicher, and N.~Navab, ``Mm-or: A large multimodal operating room dataset for semantic understanding of high-intensity surgical environments,'' in \emph{Proc. IEEE/CVF Conf. Comput. Vis. Pattern Recognit. (CVPR)}, 2025, pp. 19\,378--19\,389.

\bibitem{bib289}
N.~L. Rodas, F.~Barrera, and N.~Padoy, ``See it with your own eyes: Markerless mobile augmented reality for radiation awareness in the hybrid room,'' \emph{IEEE Trans. Biomed. Eng.}, vol.~64, no.~2, pp. 429--440, 2016.

\bibitem{bib290}
D.~Hu, S.~Li, and M.~Wang, ``Object detection in hospital facilities: A comprehensive dataset and performance evaluation,'' \emph{Eng. Appl. Artif. Intell.}, vol. 123, p. 106223, 2023.

\bibitem{bib291}
F.~S. Bashiri, E.~LaRose, P.~Peissig, and A.~P. Tafti, ``Mcindoor20000: A fully-labeled image dataset to advance indoor objects detection,'' \emph{Data Brief}, vol.~17, pp. 71--75, 2018.

\bibitem{bib292}
A.~Ismail, S.~A. Ahmad, A.~C. Soh, M.~K. Hassan, and H.~H. Harith, ``Mynursinghome: A fully-labelled image dataset for indoor object classification,'' \emph{Data Brief}, vol.~32, p. 106268, 2020.

\bibitem{bib293}
V.~Srivastav, T.~Issenhuth, A.~Kadkhodamohammadi, M.~de~Mathelin, A.~Gangi, and N.~Padoy, ``{MVOR}: A multi-view {RGB-D} operating room dataset for {2D} and {3D} human pose estimation,'' \emph{arXiv preprint arXiv:1808.08180}, 2018.

\bibitem{bib294}
K.~Chen, P.~Gabriel, A.~Alasfour, C.~Gong, W.~K. Doyle, O.~Devinsky, D.~Friedman, P.~Dugan, L.~Melloni, T.~Thesen \emph{et~al.}, ``Patient-specific pose estimation in clinical environments,'' \emph{IEEE J. Transl. Eng. Health Med.}, vol.~6, pp. 1--11, 2018.

\bibitem{bib295}
V.~Markova, T.~Ganchev, S.~Filkova, and M.~Markov, ``{MMD-MSD}: A multimodal multisensory dataset in support of research and technology development for musculoskeletal disorders,'' \emph{Algorithms}, vol.~17, no.~5, p. 187, 2024.

\bibitem{bib296}
J.~Wu, Z.~Chen, and M.~Xu, ``Surgtrack: {CAD}-free {3D} tracking of real-world surgical instruments,'' in \emph{Proc. Int. Conf. Med. Image Comput. Comput.-Assist. Interv. Workshops (MICCAI Workshops)}, vol. 15274, 2025, p. 168.

\bibitem{bib297}
M.~J. Sekiavandi, L.~Dixen, J.~Fimland, S.~K. Desu, A.-B. Zserai, Y.~S. Lee, M.~Barrett, and P.~Burelli, ``Advancing face-to-face emotion communication: A multimodal dataset ({AFFEC}),'' \emph{arXiv preprint arXiv:2504.18969}, 2025.

\bibitem{bib298}
Z.~Cheng, Z.-Q. Cheng, J.-Y. He, K.~Wang, Y.~Lin, Z.~Lian, X.~Peng, and A.~Hauptmann, ``Emotion-{LLaMA}: Multimodal emotion recognition and reasoning with instruction tuning,'' \emph{Adv. Neural Inf. Process. Syst. (NeurIPS)}, vol.~37, pp. 110\,805--110\,853, 2024.

\bibitem{bib299}
P.~Yang, N.~Liu, X.~Liu, Y.~Shu, W.~Ji, Z.~Ren, J.~Sheng, M.~Yu, R.~Yi, D.~Zhang \emph{et~al.}, ``A multimodal dataset for mixed emotion recognition,'' \emph{Sci. Data}, vol.~11, no.~1, p. 847, 2024.

\bibitem{bib300}
W.-B. Jiang, X.-H. Liu, W.-L. Zheng, and B.-L. Lu, ``Seed-{VII}: A multimodal dataset of six basic emotions with continuous labels for emotion recognition,'' \emph{IEEE Trans. Affect. Comput.}, 2024.

\bibitem{bib301}
R.~Subramanian, J.~Wache, M.~K. Abadi, R.~L. Vieriu, S.~Winkler, and N.~Sebe, ``{ASCERTAIN}: Emotion and personality recognition using commercial sensors,'' \emph{IEEE Trans. Affect. Comput.}, vol.~9, no.~2, pp. 147--160, 2016.

\bibitem{bib302}
S.~Katsigiannis and N.~Ramzan, ``{DREAMER}: A database for emotion recognition through {EEG} and {ECG} signals from wireless low-cost off-the-shelf devices,'' \emph{IEEE J. Biomed. Health Inform.}, vol.~22, no.~1, pp. 98--107, 2017.

\bibitem{bib303}
M.~Hu, P.~Xia, L.~Wang, S.~Yan, F.~Tang, Z.~Xu, Y.~Luo, K.~Song, J.~Leitner, X.~Cheng \emph{et~al.}, ``Ophnet: A large-scale video benchmark for ophthalmic surgical workflow understanding,'' in \emph{Proc. Eur. Conf. Comput. Vis. (ECCV)}, 2024, pp. 481--500.

\bibitem{bib304}
Z.~Wang, B.~Lu, Y.~Long, F.~Zhong, T.-H. Cheung, Q.~Dou, and Y.-H. Liu, ``Autolaparo: A dataset of integrated multi-tasks for image-guided surgical automation in laparoscopic hysterectomy,'' in \emph{Proc. Int. Conf. Med. Image Comput. Comput.-Assist. Interv. (MICCAI)}, 2022, pp. 486--496.

\bibitem{bib305}
A.~Derath{\'e}, F.~Reche, S.~Guy, K.~Charri{\`e}re, B.~Trilling, P.~Jannin, A.~Moreau-Gaudry, B.~Gibaud, and S.~V{\'o}r{\"o}s, ``{LapEx}: A multimodal dataset for context recognition and practice assessment in laparoscopic surgery,'' \emph{Sci. Data}, vol.~12, no.~1, p. 342, 2025.

\bibitem{bib306}
A.~Huaulm{\'e}, D.~Sarikaya, K.~Le~Mut, F.~Despinoy, Y.~Long, Q.~Dou, C.-B. Chng, W.~Lin, S.~Kondo, L.~Bravo-S{\'a}nchez \emph{et~al.}, ``Micro-surgical anastomose workflow recognition challenge report,'' \emph{Comput. Methods Programs Biomed.}, vol. 212, p. 106452, 2021.

\bibitem{bib307}
Z.~Wu, D.~Tong, H.~Xie, L.~Sun, X.~Fan, and Z.~Yang, ``A portable {6D} surgical instrument magnetic localization system with dynamic error correction,'' \emph{IEEE Sens. J.}, 2025.

\bibitem{bib308}
Z.~Qi, H.~Jin, X.~Xu, Q.~Wang, Z.~Gan, R.~Xiong, S.~Zhang, M.~Liu, J.~Wang, X.~Ding \emph{et~al.}, ``Head model dataset for mixed reality navigation in neurosurgical interventions for intracranial lesions,'' \emph{Sci. Data}, vol.~11, no.~1, p. 538, 2024.

\bibitem{bib309}
F.~Xia, A.~R. Zamir, Z.~He, A.~Sax, J.~Malik, and S.~Savarese, ``Gibson {Env}: Real-world perception for embodied agents,'' in \emph{Proc. IEEE/CVF Conf. Comput. Vis. Pattern Recognit. (CVPR)}, 2018, pp. 9068--9079.

\bibitem{bib310}
S.~K. Ramakrishnan, A.~Gokaslan, E.~Wijmans, O.~Maksymets, A.~Clegg, J.~Turner, E.~Undersander, W.~Galuba, A.~Westbury, A.~X. Chang \emph{et~al.}, ``{HM3D}: 1000 large-scale {3D} environments for embodied {AI},'' \emph{arXiv preprint arXiv:2109.08238}, 2021.

\bibitem{bib311}
K.~Yuan, M.~Kattel, J.~L. Lavanchy, N.~Navab, V.~Srivastav, and N.~Padoy, ``Advancing surgical {VQA} with scene graph knowledge,'' \emph{Int. J. Comput. Assist. Radiol. Surg.}, vol.~19, no.~7, pp. 1409--1417, 2024.

\bibitem{bib312}
S.~Ray, K.~Gupta, S.~Kundu, P.~A. Kasat, S.~Aditya, and P.~Goyal, ``{ERVQA}: A dataset to benchmark the readiness of large vision--language models in hospital environments,'' \emph{arXiv preprint arXiv:2410.06420}, 2024.

\bibitem{bib313}
J.~Wu, W.~Deng, X.~Li, S.~Liu, T.~Mi, Y.~Peng, Z.~Xu, Y.~Liu, H.~Cho, C.-I. Choi \emph{et~al.}, ``{MedReason}: Eliciting factual medical reasoning steps in {LLMs} via knowledge graphs,'' \emph{arXiv preprint arXiv:2504.00993}, 2025.

\bibitem{bib314}
Y.~Sun, X.~Qian, W.~Xu, H.~Zhang, C.~Xiao, L.~Li, D.~Zhao, W.~Huang, T.~Xu, Q.~Bai \emph{et~al.}, ``Reasonmed: A 370k multi-agent generated dataset for advancing medical reasoning,'' in \emph{Proc. Conf. Empir. Methods Nat. Lang. Process. (EMNLP)}, 2025, pp. 26\,457--26\,478.

\bibitem{bib315}
E.~{\"O}zsoy, C.~Pellegrini, D.~Bani-Harouni, K.~Yuan, M.~Keicher, and N.~Navab, ``{ORQA}: A benchmark and foundation model for holistic operating room modeling,'' \emph{arXiv preprint arXiv:2505.12890}, 2025.

\bibitem{bib316}
J.~Li, G.~Skinner, G.~Yang, B.~R. Quaranto, S.~D. Schwaitzberg, P.~C.~W. Kim, and J.~Xiong, ``{LLaVA-Surg}: Towards a multimodal surgical assistant via structured surgical video learning,'' \emph{arXiv preprint arXiv:2408.07981}, 2024.

\bibitem{bib326}
J.~Xu, B.~Li, B.~Lu, Y.-H. Liu, Q.~Dou, and P.-A. Heng, ``{SurRoL}: An open-source reinforcement learning centered and {dVRK}-compatible platform for surgical robot learning,'' in \emph{Proc. IEEE/RSJ Int. Conf. Intell. Robots Syst. (IROS)}, 2021, pp. 1821--1828.

\bibitem{bib327}
Q.~Yu, M.~Moghani, K.~Dharmarajan, V.~Schorp, W.~C.-H. Panitch, J.~Liu, K.~Hari, H.~Huang, M.~Mittal, K.~Goldberg \emph{et~al.}, ``{ORBIT-Surgical}: An open simulation framework for learning surgical augmented dexterity,'' in \emph{Proc. IEEE Int. Conf. Robot. Autom. (ICRA)}, 2024, pp. 15\,509--15\,516.

\bibitem{bib328}
S.~Schmidgall, A.~Krieger, and J.~Eshraghian, ``{Surgical Gym}: A high-performance {GPU}-based platform for reinforcement learning with surgical robots,'' in \emph{Proc. IEEE Int. Conf. Robot. Autom. (ICRA)}, 2024, pp. 13\,354--13\,361.

\bibitem{bib330}
Y.~Ao, M.~Moghani, M.~Mittal, M.~Prajapat, L.~Wu, F.~Giraud, F.~Carrillo, A.~Krause, and P.~F{\"u}rnstahl, ``{SonoGym}: High-performance simulation for challenging surgical tasks with robotic ultrasound,'' \emph{arXiv preprint arXiv:2507.01152}, 2025.

\bibitem{bib317}
Y.~Gao, S.~S. Vedula, C.~E. Reiley, N.~Ahmidi, B.~Varadarajan, H.~C. Lin, L.~Tao, L.~Zappella, B.~B{\'e}jar, D.~D. Yuh \emph{et~al.}, ``{JIGSAWS}: A surgical activity dataset for human motion modeling,'' in \emph{MICCAI Workshop on Modeling and Monitoring of Computer Assisted Interventions (M2CAI)}, 2014.

\bibitem{bib322}
R.~Stauder, D.~Ostler, M.~Kranzfelder, S.~Koller, H.~Feu{\ss}ner, and N.~Navab, ``The {TUM LapChole} dataset for the {M2CAI} 2016 workflow challenge,'' \emph{arXiv preprint arXiv:1610.09278}, 2016.

\bibitem{bib320}
J.~L. Lavanchy, S.~Ramesh, D.~Dall'Alba, C.~Gonzalez, P.~Fiorini, B.~P. M{\"u}ller-Stich, P.~C. Nett, J.~Marescaux, D.~Mutter, and N.~Padoy, ``Challenges in multi-centric generalization: Phase and step recognition in {Roux-en-Y} gastric bypass surgery,'' \emph{Int. J. Comput. Assist. Radiol. Surg.}, vol.~19, no.~11, pp. 2249--2257, 2024.

\bibitem{bib321}
A.~Zia, M.~Berniker, R.~Nespolo, C.~Perreault, Z.~Wang, B.~Mueller, R.~Schmidt, K.~Bhattacharyya, X.~Liu, and A.~Jarc, ``{SurgVU}: Surgical visual understanding dataset,'' \emph{arXiv preprint arXiv:2501.09209}, 2025.

\bibitem{bib323}
S.~Schmidgall, J.~W. Kim, J.~Jopling, and A.~Krieger, ``General surgery vision transformer: A video pre-trained foundation model for general surgery,'' \emph{arXiv preprint arXiv:2403.05949}, 2024.

\bibitem{bib324}
R.~Hartwig, D.~Ostler, J.-C. Rosenthal, H.~Feu{\ss}ner, D.~Wilhelm, and D.~Wollherr, ``{MITI}: {SLAM} benchmark for laparoscopic surgery,'' \emph{arXiv preprint arXiv:2202.11496}, 2022.

\bibitem{bib325}
G.~Wang, H.~Xiao, R.~Zhang, H.~Gao, L.~Bai, X.~Yang, Z.~Li, H.~Li, and H.~Ren, ``{CoPESD}: A multi-level surgical motion dataset for training large vision--language models to co-pilot endoscopic submucosal dissection,'' in \emph{Proc. ACM Int. Conf. Multimedia (ACM MM)}, 2025, pp. 12\,636--12\,643.

\bibitem{bib331}
F.~Shang, J.~Fu, Y.~Yang, H.~Huang, J.~Liu, and L.~Ma, ``{SynFundus-1M}: A high-quality million-scale synthetic fundus image dataset with fifteen types of annotation,'' \emph{arXiv preprint arXiv:2312.00377}, 2023.

\bibitem{bib332}
K.~Ding, M.~Zhou, H.~Wang, O.~Gevaert, D.~Metaxas, and S.~Zhang, ``A large-scale synthetic pathological dataset for deep learning-enabled segmentation of breast cancer,'' \emph{Sci. Data}, vol.~10, no.~1, p. 231, 2023.

\bibitem{bib333}
J.~Walonoski, S.~Klaus, E.~Granger, D.~Hall, A.~Gregorowicz, G.~Neyarapally, A.~Watson, and J.~Eastman, ``{Synthea}: Novel coronavirus ({COVID-19}) model and synthetic dataset,'' \emph{Intell.-Based Med.}, vol.~1, p. 100007, 2020.

\bibitem{bib334}
J.~Walonoski, D.~Hall, K.~M. Bates, M.~H. Farris, J.~Dagher, M.~E. Downs, R.~T. Sivek, B.~Wellner, A.~Gregorowicz, M.~Hadley \emph{et~al.}, ``The ``coherent dataset'': Combining patient data and imaging in a comprehensive synthetic health record,'' \emph{Electronics}, vol.~11, no.~8, p. 1199, 2022.

\bibitem{bib335}
H.~Guan and M.~Liu, ``Domain adaptation for medical image analysis: A survey,'' \emph{IEEE Trans. Biomed. Eng.}, vol.~69, no.~3, pp. 1173--1185, 2021.

\bibitem{bib336}
D.~Wang, Y.~Zhang, K.~Zhang, and L.~Wang, ``{FocalMix}: Semi-supervised learning for {3D} medical image detection,'' in \emph{Proc. IEEE/CVF Conf. Comput. Vis. Pattern Recognit. (CVPR)}, 2020, pp. 3951--3960.

\bibitem{bib337}
I.~Dayan, H.~R. Roth, A.~Zhong, A.~Harouni, A.~Gentili, A.~Z. Abidin, A.~Liu, A.~B. Costa, B.~J. Wood, C.-S. Tsai \emph{et~al.}, ``Federated learning for predicting clinical outcomes in patients with {COVID-19},'' \emph{Nat. Med.}, vol.~27, no.~10, pp. 1735--1743, 2021.

\bibitem{bib338}
Z.~Sun, H.~Yin, H.~Chen, T.~Chen, L.~Cui, and F.~Yang, ``Disease prediction via graph neural networks,'' \emph{IEEE J. Biomed. Health Inform.}, vol.~25, no.~3, pp. 818--826, 2020.

\bibitem{bib339}
H.~Wu, W.~Shi, A.~Choudhary, and M.~D. Wang, ``Clinical decision making under uncertainty: A bootstrapped counterfactual inference approach,'' \emph{BMC Med. Inform. Decis. Mak.}, vol.~24, no.~1, p. 275, 2024.

\bibitem{bib24}
M.~Sharifi, S.~Tripathi, Y.~Chen, Q.~Zhang, and M.~Tavakoli, ``Reinforcement learning methods for assistive and rehabilitation robotic systems: A survey,'' \emph{IEEE Trans. Syst., Man, Cybern., Syst.}, 2025.

\end{thebibliography}

\vfill

\end{document}